%% file: main.tex
\documentclass[10pt,twocolumn,letterpaper]{article}

\usepackage{wacv}              

\usepackage{graphicx}
\usepackage{amsmath}
\usepackage{amssymb}
\usepackage{booktabs}
\usepackage{makecell}
\usepackage{floatrow}
\usepackage{multirow}
\usepackage[page,header]{appendix}
\usepackage{titletoc}
\usepackage[pagebackref,breaklinks,colorlinks]{hyperref}

\usepackage[capitalize]{cleveref}
\crefname{section}{Sec.}{Secs.}
\Crefname{section}{Section}{Sections}
\Crefname{table}{Table}{Tables}
\crefname{table}{Tab.}{Tabs.}

\begin{document}
\title{Street TryOn: Learning In-the-Wild Virtual Try-On from Unpaired Person Images}

\author{Aiyu Cui \quad Jay Mahajan \quad Viraj Shah \quad Preeti Gomathinayagam \quad Chang Liu \quad Svetlana Lazebnik \\
University of Illinois Urbana-Champaign \\
{\url{https://cuiaiyu.github.io/StreetTryOn}}
}

\twocolumn[{%
\renewcommand\twocolumn[1][]{#1}%
\maketitle
\begin{center}
\vspace{-10mm}
    \centering
    \includegraphics[width=\textwidth]{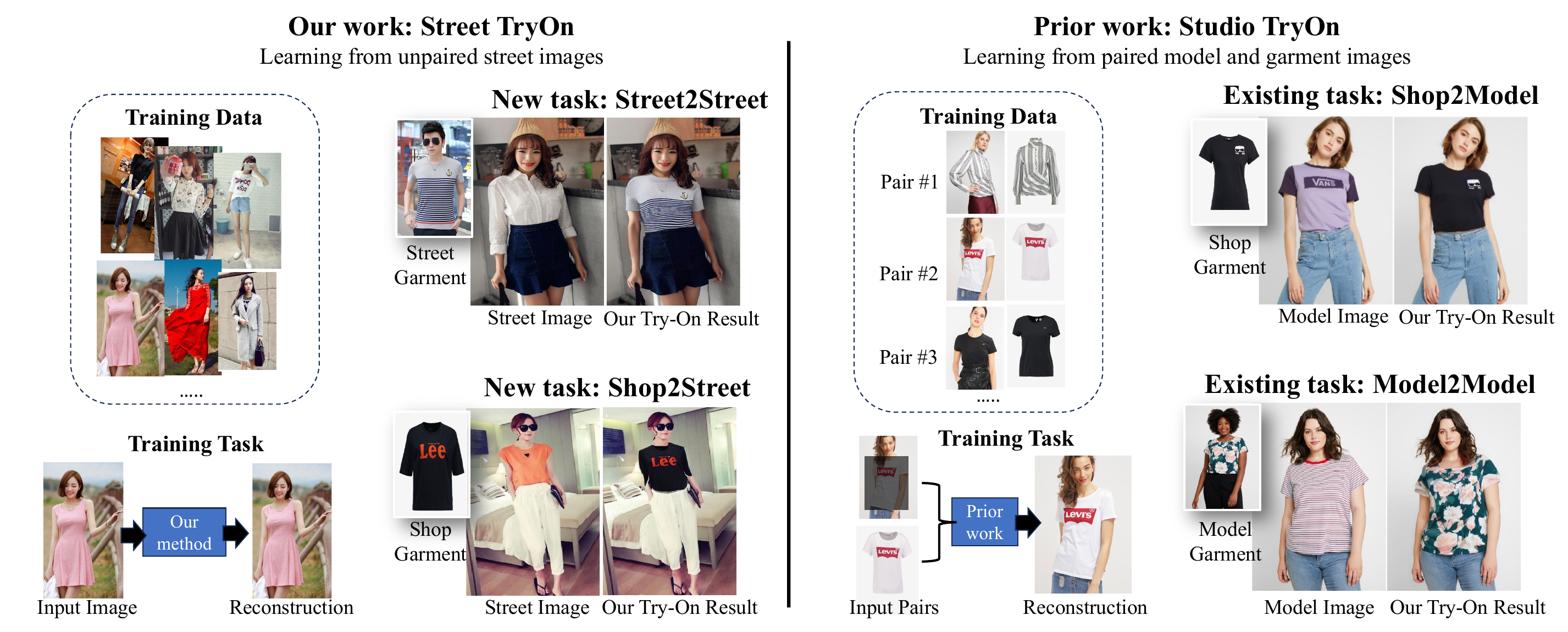}
    \vspace{-8mm}
    \captionof{figure}{Our proposed Street TryOn benchmark and method for \textbf{in-the-wild try-on} application, contrasted with existing work focusing on controlled studio images and paired training (see text).}
    \label{fig:teaser}
    
    \vspace{-2mm}
\end{center}%
}]
\input{sec/0_abstract}    
\input{sec/1_intro}

\input{sec/2_related_work}

\input{sec/3_dataset}
\input{sec/4_methods}

\input{sec/5_experiments}
\input{sec/6_conclusion}
{\small
\bibliographystyle{ieee_fullname}
\bibliography{egbib}
}

\newpage
\appendix
\appendixpage
\startcontents[sections]
\printcontents[sections]{l}{1}{\setcounter{tocdepth}{2}}
\input{supp/dataset_filtering_details}

\input{supp/cosine_noise_details}

\input{supp/diffusion_inpainting_details}
\newpage
\input{supp/more_results}

\end{document}

%% file: sec/0_abstract.tex
\begin{abstract}
\vspace{-5mm}
Most virtual try-on research is motivated to serve the fashion business by generating images to demonstrate garments on studio models at a lower cost. However, virtual try-on should be a broader application that also allows customers to visualize garments on themselves using their own casual photos, known as \textbf{in-the-wild try-on}. Unfortunately, the existing methods, which achieve plausible results for studio try-on settings, perform poorly in the in-the-wild context. This is because these methods often require paired images (garment images paired with images of people wearing the same garment) for training. While such paired data is easy to collect from shopping websites for studio settings, it is difficult to obtain for in-the-wild scenes.

In this work, we fill the gap by (1) introducing a \textbf{StreetTryOn} benchmark to support in-the-wild virtual try-on applications and (2) proposing a novel method to learn virtual try-on from a set of in-the-wild person images directly without requiring paired data. We tackle the unique challenges, including warping garments to more diverse human poses and rendering more complex backgrounds faithfully, by a novel DensePose warping correction method combined with diffusion-based conditional inpainting. Our experiments show competitive performance for standard studio try-on tasks and SOTA performance for street try-on and cross-domain try-on tasks.

\end{abstract}

%% file: sec/1_intro.tex
\vspace{-5mm}
\section{Introduction}
\vspace{-2mm}
\label{sec:intro}
 Driven by the fashion industry's demand, image-based virtual try-on have rapidly become a popular research topic.  Most existing research focuses on serving fashion merchants by using AI to generate images of models wearing different garments, offering a cost-effective alternative to hiring models for physical photoshoots in the studio. As a result, virtual try-on research has achieved high levels of performance in transferring in-shop garments to model images or from one model image to another~\cite{ge2021pfafn,he2022styleflow,xie2023gp-vton,zhu2023tryondiffusion,albahar2021posewithstyle,cui2021dior,xie2021pastagan,xie2022pastagan++} (Fig.~\ref{fig:teaser}, right). 
 However, in this work, we target a rarely studied virtual try-on application, \textbf{in-the-wild try-on}, which is to transfer garments to and from in-the-wild images (Fig.~\ref{fig:teaser}, left). This application allows the general population to visualize how a garment would look on their own bodies by taking photos casually. It has significant commercial potential and will be a key research direction for virtual try-on as advanced image generation models, like diffusion models, become available.

Virtual try-on tasks generally face two key challenges: warping a garment to fit a person's body and fusing the warped garment faithfully with the person's image. In-the-wild try-on intensifies these challenges because it must handle a more diverse range of ``street'' data to fit garments to more casual poses and camera angles, as well as harmonize images with more complex backgrounds and lighting conditions. Existing studio try-on methods~\cite{ge2021pfafn,he2022styleflow,xie2023gp-vton,zhu2023tryondiffusion,albahar2021posewithstyle,cui2021dior,kim2023stableviton}, trained on model images with standard poses, white backgrounds, and uniform lighting, cannot be directly applied to in-the-wild try-on due to the domain shift. As we will demonstrate, these methods struggle with limb reconstruction, warping, and background rendering. Additionally, the existing studio try-on methods require paired images from shopping websites to learn garment warping (as shown in Fig.~\ref{fig:teaser}, right), but such paired data is hard to collect for in-the-wild try-on settings. Consequently, to enable in-the-wild try-on, we need to solve the more challenging warping and fusion problems with mostly unpaired training data.

Since there is no formal dataset to evaluate the in-the-wild virtual try-on, we first introduce a new benchmark, \textbf{StreetTryOn}. StreetTryOn is derived from the large fashion retrieval dataset DeepFashion2~\cite{ge2019deepfashion2} by filtering out over the images that are infeasible for try-on tasks (e.g., non-frontal view, large occlusion, dark environment, etc.), resulting in $12K$ training and $2K$ test images.
Combining with the images in VITON-HD dataset~\cite{choi2021vitonhd}, we establish a suite of try-on tasks that have garment and person inputs from various sources. Of primary interest to us are the new tasks of Shop2Street and Street2Street garment transfer (Fig.~\ref{fig:teaser}, left), but for completeness, we also include the more traditional tasks of Shop2Model and Model2Model (Fig.~\ref{fig:teaser}, right). Benchmarking methods across all these tasks can give a comprehensive idea of the robustness and cross-domain generalization ability of different approaches (i.e., generalization of methods trained on ``studio'' images to ``street'' images, and vice versa). 

\begin{figure*}
    \centering
    \vspace{-5mm}
    \includegraphics[width=0.75\textwidth]{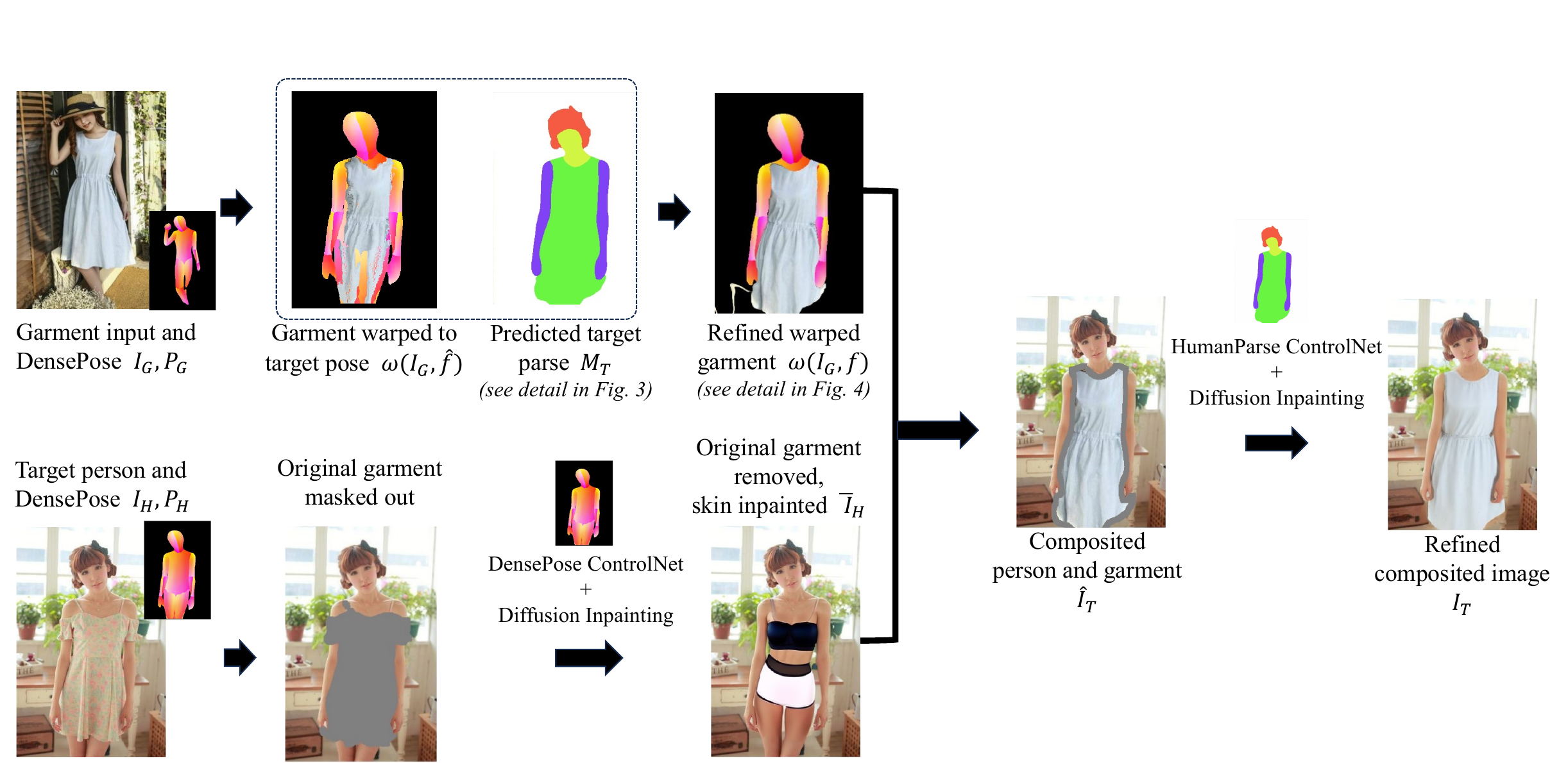}
    \caption{Overview of our proposed virtual try-on method (see text for details).
    \label{fig:overview}}
    \vspace{-2mm}
\end{figure*}

To obtain robust performance on the in-the-wild try-on application (the Shop2Street and Street2Street tasks), in Sec.~\ref{sec:methods}, we introduce a novel approach for learning virtual try-on from unpaired in-the-wild person images with an
overview shown in Fig.~\ref{fig:overview}.
Because the amount of StreetTryOn data is insufficient for training a high-quality model from scratch, we leverage several powerful and robust pre-trained components, most notably, DensePose~\cite{guler2018densepose} correspondence to perform garment warping, and diffusion model inpainting to remove the old garment, inpaint skin, and composite the warped target garment onto the person. 
These components are directly dictated by the challenge of learning from unpaired street images. Without paired training data, it is not possible to directly learn a 2D warping function from source to target images. Instead, DensePose allows us to obtain a warp estimate by registration to a common 3D model. Our warping correction module refines this estimate, with novel training-time data augmentation to mimic registration errors (Sec.~\ref{sec:warp}). Our skin inpainting module addresses the limb reconstruction challenge by leveraging the pre-trained strong diffusion inpainter.

Comprehensive evaluation in Sec. \ref{sec:exp} will show that our method outperforms all existing methods on our StreetTryOn benchmark and is competitive on the much more mature VITON-HD benchmark~\cite{choi2021vitonhd}. Our method is remarkably robust for the hardest try-on setting, Street2Street, achieving similar results whether trained on in-domain or out-of-domain data.





%% file: sec/2_related_work.tex
\section{Related Work}
\label{sec:related_work}

\noindent \textbf{Virtual Try-On Benchmarks.}
Existing Shop2Model virtual try-on benchmarks include VITON~\cite{han2018viton}, VITON-HD~\cite{choi2021vitonhd}, MPV~\cite{dong2019mpv}, and DressCode~\cite{morelli2022dresscode}, all of which have paired person and garment images with studio model as person source and ghost mannequin images as garment source. VITON, VITON-HD, and MPV only have top garments, but DressCode also includes other categories like pants and dresses. MPV has multiple person images paired with the garment images in different poses. DeepFashion\cite{liu2016deepfashion}, and UPT~\cite{xie2022pastagan++} datasets have also been used to demonstrate Model2Model try-on. However, none of the existing datasets are representative of in-the-wild try-on settings. Therefore, our proposed StreetTryOn benchmark is a necessary addition to the literature.
The SHHQ-1.0 dataset~\cite{fu2022shhq} has previously been proposed to evaluate in-the-wild try-on performance. However, at least 25\% images in SHHQ-1.0 are studio model images aggregated from the DeepFashion dataset~\cite{liu2016deepfashion} and the African fashion ~dataset\cite{hacheme2021africanfashion}. 

\noindent \textbf{Virtual Try-On Approaches.} Most of the top-performing methods for the Shop2Model try-on~\cite{ge2021pfafn,han2019clothflow, he2022styleflow,xie2023gp-vton, zhu2023tryondiffusion} are trained on paired datasets mentioned above, like VITON-HD~\cite{choi2021vitonhd} and DressCode~\cite{morelli2022dresscode}. Such methods can achieve high-quality results on in-domain images, but do not transfer well to in-the-wild data. 
Several other works~\cite{albahar2021posewithstyle, cui2021dior, zhu2023tryondiffusion} can achieve Model2Model try-on by training on paired data (people wearing the same outfits in multiple poses). However, such methods cannot be trained in settings where paired data is unavailable.
PASTAGAN~\cite{xie2021pastagan} and PASTAGAN++~\cite{xie2022pastagan++} are the only prior works for Model2Model try-on trained without paired training data on the UPT dataset~\cite{xie2021pastagan}. However, our experiments will show that PASTAGAN++ cannot handle the complex backgrounds of street images. 
Dressing-in-the-wild \cite{dong2022dressing-in-the-wild} is another method aimed at in-the-wild try-on, but both its training and inference require videos to learn the person representation.

To capture the complicated appearancce distribution of dressed people, a few prior works~\cite{men2020controllable,lewis2021tryongan} have used a StyleGAN-like architecture~\cite{karras2019stylegan} with encoded garments as style codes, but such codes typically have a hard time preserving garment details. PASTAGAN++~\cite{xie2022pastagan++}, one of the few previous methods that can be trained from unpaired data, introduces a sophisticated patch-based garment representation to effectively preserve details. However, its StyleGAN nature makes it fundamentally difficult to train on street images with varied backgrounds.

Previous methods usually learn warping from paired data. Clothflow~\cite{han2019clothflow} uses a pyramid architecture to predict flow fields. PFAFN~\cite{ge2021pfafn} learns flow from the correlation between garment and person; FS-VTON~\cite{he2022styleflow} leverages StyleGAN to predict flow; GP-VTON~\cite{xie2023gp-vton} warps garments part by part with multiple local flows.
TryOnDiffusion~\cite{zhu2023tryondiffusion} learns implicit flows via cross-attention from paired data. Like our approach, Pose-with-style~\cite{albahar2021posewithstyle} also uses DensePose to perform warping, but it has to learn from paired data to refine the warping and rendering.

\noindent \textbf{Diffusion Models}~\cite{saharia2022imagen,ramesh2022dalle, rombach2022stablediffusion} have shown a seemingly magical power of image generation. In particular, one can arbitrarily change people's outfits using text-guided diffusion inpainting. However, this does not provide sufficient control for the virtual try-on task, which requires faithful transfer of a specific garment from another image. To our knowledge, TryOnDiffusion~\cite{zhu2023tryondiffusion} is a specialized diffusion model for virtual try-on so far, trained from scratch on a proprietary dataset of millions of model image pairs. This work has not been released, so we cannot compare with it. By contrast, our method leverages the power of pre-trained diffusion inpainting and only requires the training of a few lightweight additional components, which can be done on a relatively small amount of unpaired data.
StableVITON~\cite{kim2023stableviton} is also a specialized diffusion model that warps garments implicitly using cross-attention in the U-Net, which has to be learned from paired training data. Although StableVITON shows SOTA performance on Shop2Model try-on and promising results for street try-on, it still struggles with garment detail preservation, limb reconstruction, and background rendering. Plus, StableVITON only supports taking garments from the ghost mannequin images, while our methods can allow garments from either shop images or from a person. 

Several pretrained diffusion models for text-to-image generation have been released in the past year~\cite{saharia2022imagen,ramesh2022dalle, rombach2022stablediffusion}. Stable Diffusion~\cite{rombach2022stablediffusion} is one of the most accessible because of its open-source code and model weights. 
It supports inpainting, where one can mask a certain area in an image and fill it in guided by a text prompt. 
ControlNet~\cite{zhang2023controlnet} enables tighter conditioning of pre-trained diffusion models using additional inputs like pose and semantic segmentation. We train two DensePose-conditioned ControlNets to enable the removal of the old garment and compositing of the new garment warped to the target pose.

%% file: sec/3_dataset.tex
\section{StreetTryOn Benchmark}
\label{sec:benchmark}
To explore in-the-wild and cross-domain try-on, we introduce a new benchmark called \textbf{StreetTryOn}, derived from the existing fashion retrieval dataset DeepFashion2~\cite{ge2019deepfashion2}. DeepFashion2 contains $191,961$ training and $32,153$ test images of people with diverse garments and backgrounds, but unfortunately, most of them cannot directly be used for virtual try-on since they only show portions of the body, have large occlusions, non-frontal views, or dark lighting conditions.
To remove such unsuitable images, we apply a multi-step filtering process using a combination of provided DeepFashion2 annotations, person detection, and manual selection, resulting in a clean set of $12,364$ training and $2,089$ test images. More details about the filtering are provided in the supplementary materials. 


\noindent\textbf{Benchmark Tasks.} The try-on tasks of greatest interest to us are \textbf{Street2Street}, \textbf{Shop2Street}, and \textbf{Model2Street} (Fig.~\ref{fig:teaser}). For the latter two cross-domain tasks, we obtain the needed shop and model test images from VITON-HD~\cite{choi2021vitonhd}. For \textbf{Street2Street}, we use the $2,089$ test street images in StreetTryOn, which are partitioned into two subsets of $909$ ``top'' images and $1,190$ ``dresses.'' Then we construct $909$ and $1,190$ unpaired (person, garment) test tuples by random shuffling. 
For \textbf{Shop2Street} and \textbf{Model2Street} try-on, we randomly sample $909$ garment ghost mannequin images and $909$ model images from VITON-HD to construct two sets of $909$ cross-domain (person, garment) test tuples. 
Combining the above test sets with existing \textbf{Shop2Model} and \textbf{Model2Model} test sets from VITON-HD gives us a comprehensive suite of scenarios for evaluation.

%% file: sec/4_methods.tex
\section{Our Try-On Method}
\label{sec:methods}
An overview of our method is shown in Fig.~\ref{fig:overview}. Given a person image $I_H$ and a garment image $I_G$\footnote{For simplicity, we use $I_G$ to denote the garment image with everything except for the try-on garment masked out.}, our goal is to generate the try-on image $I_T$ with person $I_H$ wearing $I_G$. We preprocess $I_H$ and $I_G$ to obtain human parses $M_H$ and $M_G$, as well as DensePose~\cite{guler2018densepose} estimates $P_H$ and $P_G$. 



Our try-on inference pipeline starts by predicting the semantic parse $M_T$ for the try-on output image using a \textbf{TryOn Parse Estimator} (Section \ref{sec:parse}). 
Next, we predict a flow field $f$ to warp the garment $I_G$ to the output pose $P_H$ using DensePose correspondence followed by a trained \textbf{Warping Correction Module} (Section \ref{sec:warp}).
At the same time, for the person image $I_H$, we remove the original garment and inpaint skin regions by a pre-trained diffusion inpainter with a \textbf{DensePose ControlNet} conditioned on $P_H$ (Section \ref{sec:inpaint}). 
Then, we combine the warped garment $\omega(I_G, f)$ and the inpainted person $\bar{I}_H$ to get the composited person $I'_T$. Finally, we use the pre-trained diffusion inpainter with a \textbf{Human Parser ControlNet} conditioned on $M_T$ to inpaint a masked garment boundary to get the final try-on output $I_T$.

To summarize, besides the pre-trained DensePose estimator and diffusion inpainter, our method has four learnable components: TryOn Parse Estimator, Warping Correction Module, and the two ControlNets, which are all trained separately without paired data. 
In the rest of the section, we will describe each of these components in detail.

\subsection{TryOn Parse Estimator} \label{sec:parse}

\begin{figure*}[htb]
        \includegraphics[width=0.49\textwidth]{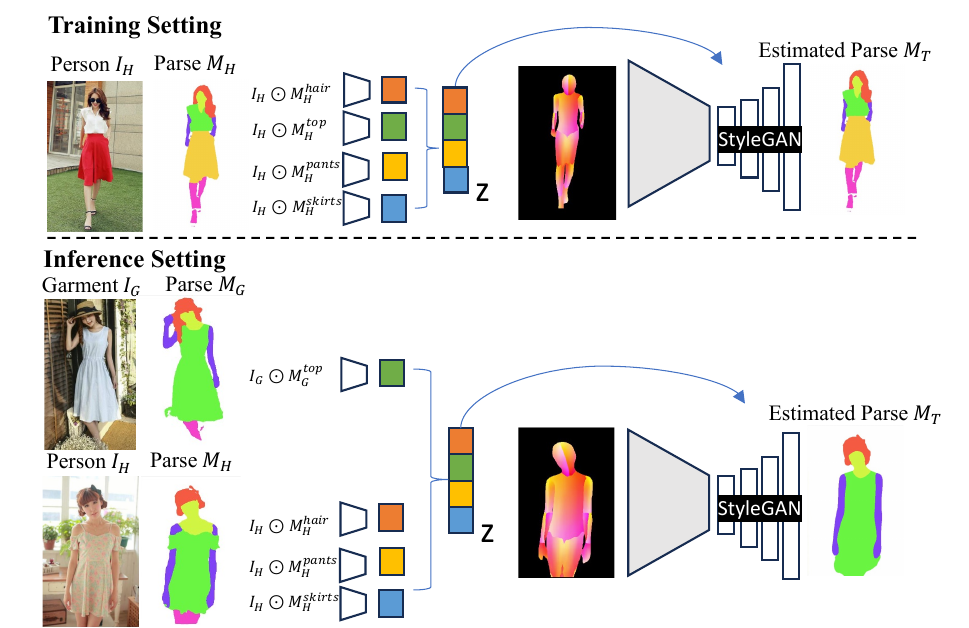}
        \hfill
    \includegraphics[width=0.49\textwidth]{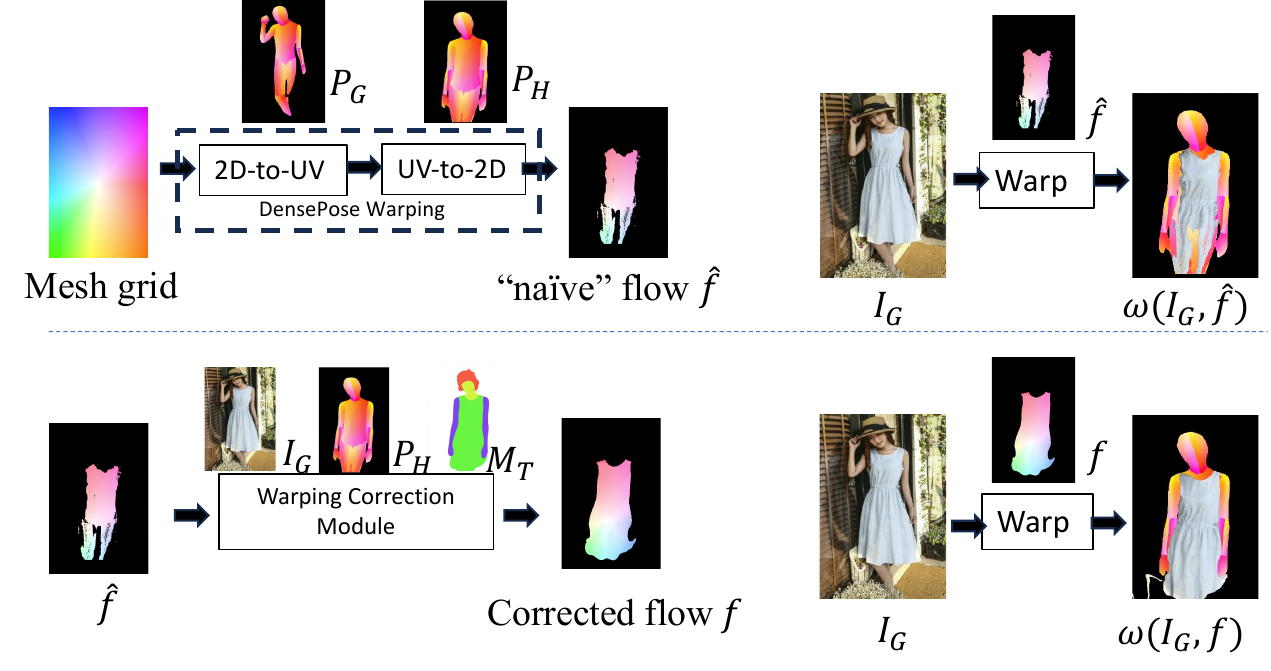}
    \vspace{-3mm}
    \caption{
            Left: TryOn Parse Estimator (Section~\ref{sec:parse}). Right: Warping Correction Module (Section~\ref{sec:warp}).
        }
    \label{fig:submodule}
\end{figure*}
 
Our try-on parse estimator has a StyleGAN architecture as shown in Fig.~\ref{fig:submodule}-left. The initial feature map of the StyleGAN $\mathbf{G}$ is the DensePose of the input person downsampled to $16\times16$ by an encoder $\mathbf{E}_{dp}$. The style code $z$ is a concatenation of four segment style codes $\{z^{top}, z^{hair}, z^{pants}, z^{skirt}\}$, encoded from the input person's outfits. Each segment $i \in \{top, hair, pants, skirt\}$ will be encoded by a segment encoder $\mathbf{E}_{seg}$ as $z^i = \mathbf{E}(I \odot M^i)$ where the mask $M_i$ of the segment $i$ is from the human parse $M$. At inference time, the top segment will come from the garment image for trying on, and the rest will come from the person input, so we predict the human parse as
\begin{equation}
    M_T = \mathbf{G}(\{z_G^{top}, z_H^{hair}, z_H^{pants}, z_H^{skirt}\} | \mathbf{E}_{dp}(P_H)).
\end{equation}

The skin regions (face, arms, legs) are excluded from $z$ to avoid the unduly bias leaked by the shape of the initially visible skin regions. For example, if the original garment of the person has long sleeves, the arm segment will only have hands, which leaks information about the sleeves. Even if the top segment is swapped with a new one with short sleeves, the model will still be biased by the hand segment to predict a long-sleeve garment. 
 
During training, all segment codes will come from the same person image, and the model is trained to reconstruct the original human parse using cross-entropy loss. We augment the training data by randomly masking the person $I_H$ to mimic test-time occlusion and applying random sheer, scaling, rotation and translation on each extracted segment $I_H\odot M_H^i$ to teach the model to handle test cases where the garments are not perfectly aligned. 

\subsection{Warping Correction Module} \label{sec:warp}

DensePose is a mapping from a person image to the coordinate system (UV space) of a parametrized 3D human model. In principle, having DensePose estimates for the person and garment image gives us a dense registration between the two, allowing us to transfer garment pixels onto the 3D model and back onto the target human in 2D. However, in practice, there are problems with this approach. DensePose estimates are far from perfect, especially for loose garments, and direct warping also results in many unwrapped areas. Therefore, we follow the DensePose warping step by a trained correction. 

As shown in the top of Fig.~\ref{fig:submodule}-right, we obtain an initial, or ``naive'' flow field $\hat{f}$ by projecting a mesh grid to the UV space using the garment's DensePose $P_G$, and then warping it back to the person's pose in image space via $P_H$.
Next, we train a correction module that takes in the naive flow $\hat{f}$ and gradually adjusts it to obtain the final flow 
\begin{equation}
    f = \mathbf{C}(\hat{f} | I_G, P_H, M_T).
\end{equation}

To train the correction module without paired data, we attempt to reconstruct the person image $I_H$ from a perturbed version $\tilde{I}_H$. First, we apply a cosine perturbation to the pixel values of DensePose $P_H$, which mimics imperfect registration at inference time (see details in the supplementary). Although this changes the the UV coordinates of $P_H$, it is still spatially aligned with the original person image $I_H$. Next, the cosine-perturbed $\tilde{P}_H$ and the image $I_H$ are spatially transformed together using a random affine transformation, translation and rotation to get perturbed versions $\tilde{P}_H$ and $\tilde{I}_H$.
Given this synthetic data, we train the corrector $\mathbf{C}$ with the same objectives as in prior work~\cite{ge2021pfafn, he2022styleflow}:
\begin{multline}
\mathcal{L} = \mathcal{L}_{smo}(\mathbf{C} (\hat{f}|\tilde{I}_H,P_H,M_H)) \\
+ \|I_H - \omega(\tilde{I}_H, \mathbf{C}(\hat{f}|\tilde{I}_H,P_H,M_H))\| \\ 
+ \mathcal{L}_{vgg} \left( I_H, \omega(\tilde{I}_H, \mathbf{C} (\hat{f}|\tilde{I}_H,P_H,M_H))\right),
\end{multline}
where $\hat{f}$ is the naive flow from $\tilde{P}_H$ to$P_H$,
$\omega(\cdot,\cdot)$ is a warping operator, 
$\mathcal{L}_{smo}$ is total variation loss for smoothing the predicted flow and $\mathcal{L}_{vgg}$ is L1 loss between VGG features~\cite{johnson2016perceptual}.

\subsection{Inpainting}
As explained at the beginning of this section, our pipeline has two inpainting steps: the first is removing the original garment and inpainting skin to get the ``undressed'' person image $\bar{I}_H$; the second is refining the composited image $\hat{I}_T$ to get the final try-on output $I_T$. 

\noindent\textbf{Garment Removal and Skin Inpainting.} \label{sec:inpaint}
To prevent information leakage from the mask used to remove the old garment, prior works erase a larger area than necessary (i.e., creating so-called ``agnostic'' person images~\cite{lee2022hrvton,bai2022sdafn,xie2023gp-vton,zhu2023tryondiffusion}) and inpaint the missing skin simultaneously with rendering the warped try-on garment. However, such an approach still struggles with generating limbs in complex poses or removing long or loose sleeves. Moreover, unlike prior works, which only deal with clean backgrounds, we want to preserve as much of the original background as possible. Therefore, 
before rendering the new garment, we introduce a separate step of removing the original garment and inpainting it with as much skin as possible. To do this by text-guided diffusion inpainting, we use the prompt ``a person in a black+ strapless++ bra++++'', where the ``+'' is used for prompt weighting~\cite{prompt-weighting}. 

\noindent \textbf{Refining the composited image.} To create the composite image $\hat{I}_T$, we take the predicted try-on garment mask, say $M_T^{top}$, out of the predicted parse $M_T$ and combine it with the ``undressed'' person image $\bar{I}_H$ as 
\begin{equation}
    \hat{I}_T = \bar{I}_H \odot e[1-M_T^{top}] + \omega(I_G, f) \odot e[M_T^{top}],
\end{equation}
where $e[\cdot]$ is an erosion operator to create a masked region along the garment boundary for the inpainter to refine, since this is the source of many errors and misalignments between the body and the warped garment tend to concentrate. Because the diffusion inpainter also tends to introduce minor changes in the image outside the inpainting mask, we copy and paste the person's face from the original person image $I_H$ to avoid distortion.

Both of the above steps are accomplished by a pre-trained Stable Diffusion inpainter~\cite{stable-diffusion-inpaint} combined with ControlNets~\cite{zhang2023controlnet} trained on our own data. Specifically, for skin inpainting, we train a ControlNet using DensePose $P_H$ as conditioning information, and for the final compositing, we train a ControlNet with predicted parse $M_T$ as conditioning. Additional implementation details are given in Section \ref{sec:exp} and the supplementary material.



%% file: sec/5_experiments.tex
\section{Experiments}
\label{sec:exp}
\begin{figure*} 
    \centering
    \includegraphics[width=0.8\textwidth]{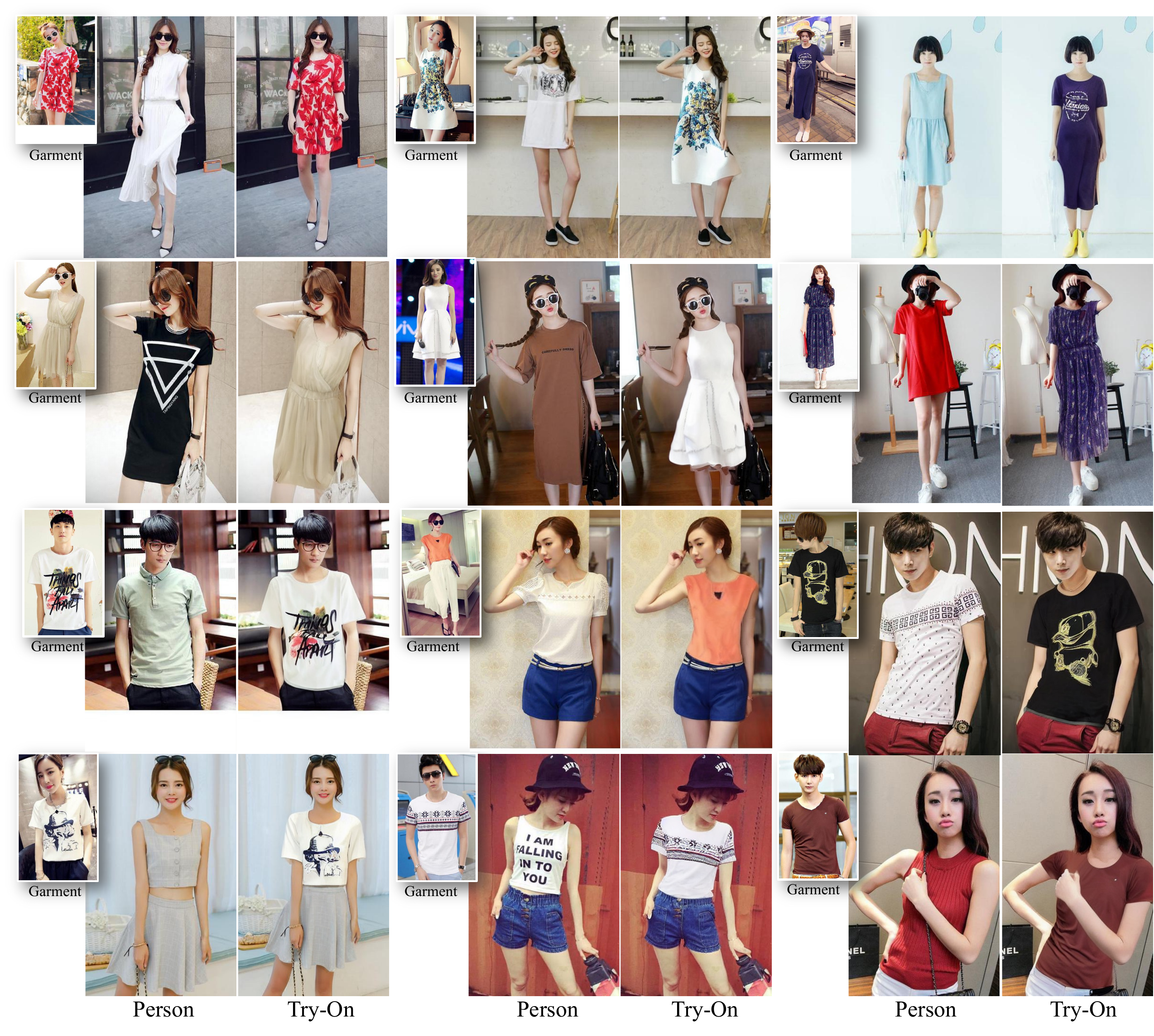}
    \vspace{-5mm}
    \caption{Street2Street Try-On examples for our method.}
    \label{fig:result}
\end{figure*}

\noindent \textbf{Evaluation Metrics.} For all tests, we evaluate performance with FID metric~\cite{heusel2017fid}, which measures the similarity of the output data distribution with the target data. We also report SSIM \cite{wang2004ssim} and LPIPS \cite{zhang2018lpips} for image reconstruction quality when paired test tuples are available. 

\noindent \textbf{Implementation Details.} 
We run experiments at $512\times 320$ resolution when the target person is from StreetTryOn (Shop2Street, Model2Street, and Street2Street tests) and $512\times 384$ resolution when the target person is from VITON-HD (Shop2Model and Model2Model).

We obtain human DensePose using the method of Guler et al.~\cite{guler2018densepose}, garment DensePose using the reimplementation  of Cui et al.~\cite{cui2023learning}, and human parse using SCHP~\cite{li2020schp}.
Our TryOn Parse Estimator is trained with a learning rate $lr=0.0001$ and batch size 8 for 100 epochs. 
Our Warping Correction module has the same architecture as the one in Clothflow~\cite{han2019clothflow} with the DensePose-warped flow as the initial input. We train it with $lr=0.00001$ for 100 epochs. For inpainting, we use the Stable Diffusion inpainter from the diffusers library~\cite{radford2021clip} and train the ControlNets in the default setting with an empty prompt for 50 epochs.

\input{tables/person2person_table_simplified}
\vspace{-5mm}
\subsection{Comparative Evaluation}

Next, we present our main experimental results. To evaluate the power of unpaired training on in-the-wild data, we compare three training settings for our method: (1) training with the standard paired VITON-HD training data; (2) unpaired training with VITON-HD person images only; (3) unpaired training with Street TryOn person images.

\vspace{-2mm}

\subsubsection{Garment Transfer from Person Images.}
\vspace{-2mm}
Tab.~\ref{tab:person2person_table} reports results for in-the-wild try-on task Street2Street, cross-domain task Model2Street, and studio task Model2Model. All of these take garments from person images (either studio or street). 
For all these settings, it is interesting to see that the training regime makes little difference for our method. This validates the robustness of all the major components of our method: TryOn Parse Estimator, DensePose warping, and ControlNet Inpainting. Fig.~\ref{fig:result} shows qualitative examples of our Street2Street results, further confirming its strong performance across poses, garment styles, and cluttered backgrounds.

\begin{figure*}
    \vspace{-3mm}

    \includegraphics[width=\textwidth]{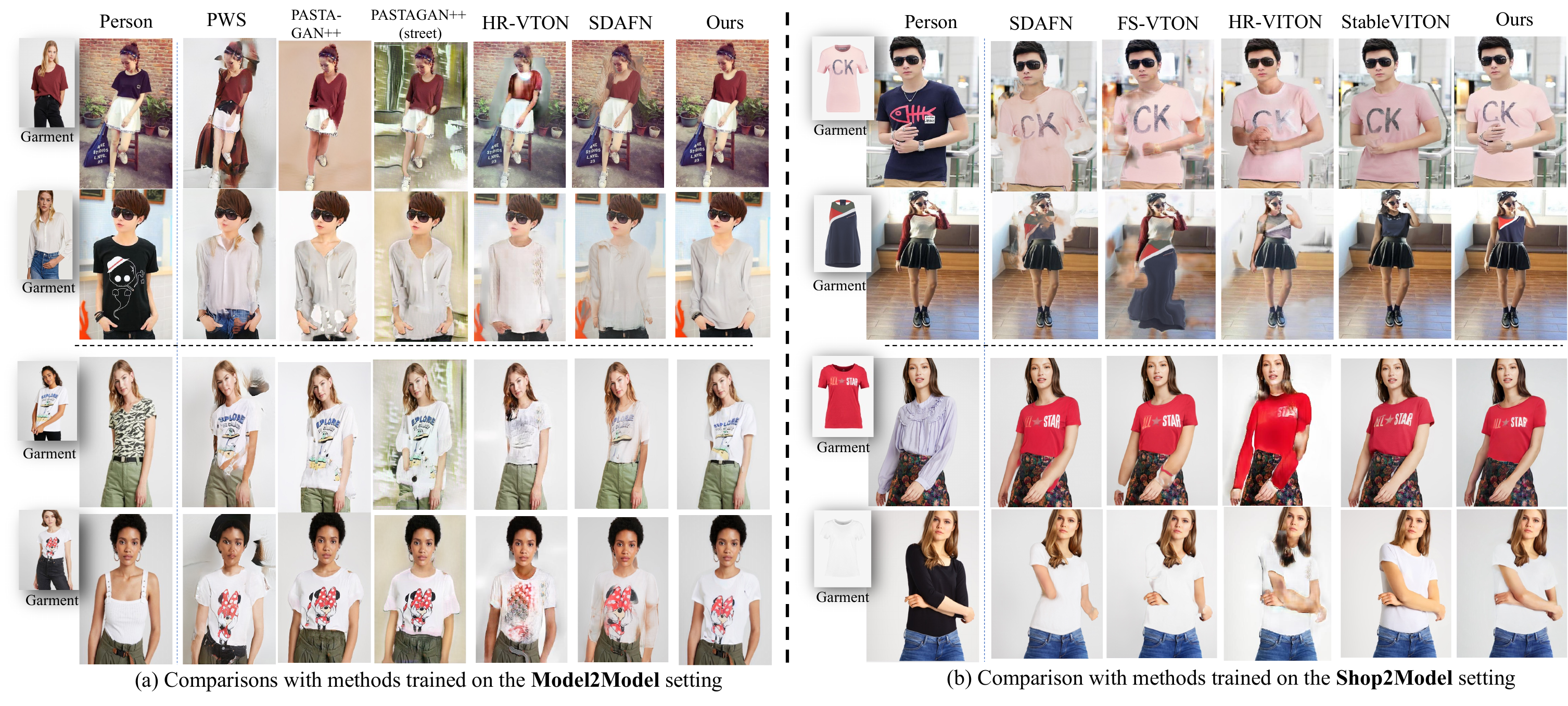}
\caption{(a)-top: Model2Street. (a)-bottom: Model2Model. (b)-top: Shop2Street. (b)-bottom: Shop2Model.}
\label{fig:main_comparison}
\vspace{-5mm}
\end{figure*}

In Tab.~\ref{tab:person2person_table}, we compare performance with Shop2Model methods FS-VTON, SDAFN and HR-VTON, which we retrain for Model2Model try-on using the DeepFashion dataset~\cite{liu2016deepfashion}. We also compare with Pose-with-Style (PWS) and PASTAGAN++, both of which are StyleGANs with customized warping designs trained in a model-to-model setting with clean backgrounds. 
In addition, we train PASTAGAN++ from scratch with the street images in the proposed Street TryOn benchmark, reported as PASTAGAN++ (street). 
In terms of FID, these methods do much worse than ours on street images, and 
Fig.~\ref{fig:main_comparison} shows why: they frequently produce warping artifacts and cannot cope with complex backgrounds. One might wonder whether the higher FIDs of competing methods are primarily due to their failure to render the background. However, in Section \ref{sec:ablation} (Fig.~\ref{fig:bg_removal}), we will present an ablation study demonstrating that even if the background is not taken into account, our method still achieves the best performance.

\input{tables/shop2person_table_simplified}

\input{tables/ablation_bg_removal}

\input{tables/ablation_effectiveness}

\vspace{-5mm}
\subsubsection{Garment Transfer from Shop Images.}
\vspace{-2mm}
Tab. \ref{tab:shop2person_table} presents an evaluation on Shop2Street and Shop2Model tasks, 
in which a garment from a ghost mannequin image is transferred to a person image. 
Here, we compare performance to SOTA Shop2Model methods PBAFN~\cite{ge2021pfafn}, FS-VTON~\cite{he2022styleflow}, SDAFN~\cite{bai2022sdafn}, GP-VTON~\cite{xie2023gp-vton},
and StableVITON~\cite{kim2023stableviton}.
Our method gets the best performance on the cross-domain Shop2Street task, confirming both its robustness and the challenging nature of our Street TryOn benchmark. Fig. \ref{fig:main_comparison} compares some qualitative results for Shop2Street. 
For most prior methods, both the warping and rendering steps tend to fail on out-of-domain street images, which is also evident in the concurrent work, StableVITON~\cite{kim2023stableviton}.
Although StableVITON shows significant improvement compared to other prior work on Shop2Street try-on, it still struggles in limb reconstruction and background rendering (Example 1 in Fig.\ref{fig:main_comparison}, right) because its agnostic person inputs tend to erase too much skin/background area. Instead, our proposed garment-removal and skin-inpainting module can better reconstruct the in-the-wild person input.
In addition, the garment reconstruction of StableVITON is not as robust as ours, because it is trained on paired studio images, which fail to fully capture the diverse distribution of in-the-wild person images. 

The last three columns of Tab. \ref{tab:shop2person_table} present a comparative evaluation on the Shop2Model (VITON-HD) task. The competing methods, which are highly tuned on VITON-HD paired data, obtain FID, SSIM, and LPIPS scores that are very close together, suggesting that the VITON-HD benchmark is nearing saturation. Our method comes within striking distance, but does not reach quite the same level quantitatively. While we train each component of our method separately, the other works jointly train the warping and refinement modules so they can better learn to complement one another. They also tend to train their flow prediction from scratch on the target dataset instead of relying on off-the-shelf DensePose. Such custom-trained flow modules give more accurate in-domain results but do not generalize well to out-of-dataset test images as we have seen. Additionally, some of these methods~\cite{ge2021pfafn, he2022styleflow, xie2023gp-vton} include a custom blending mask to eliminate the back of the garment visible in the neck area of the ghost mannequin image. Fig.~\ref{fig:main_comparison} shows qualitative examples of Shop2Model try-on. 
While our method sometimes suffers from artifacts, it has its advantages as well: it better preserves the arms, especially for hard poses, or where long sleeves need to be replaced with short ones.

\vspace{-2mm}
\subsection{Ablation Study} \label{sec:ablation}

\subsubsection{Background rendering is not the only challenge.}
\vspace{-2mm}
A reasonable inquiry is whether the main reason that studio-trained methods get bad FID scores on street images is their failure to render the complex background. Fig.~\ref{fig:bg_removal} presents an ablation study that factors out the effect of the background. To do this, we run Model2Model methods on the segmented foreground person and blend the result with the original background that has the missing area inpainted by Stable Diffusion inpainter. The table in Fig.~\ref{fig:bg_removal} compares the FID scores for three SOTA Model2Model methods with ours in three settings: foreground only, foreground blended with inpainted background, and end-to-end (inference without foreground segmentation). We can see that even the best prior method (SDAFN retrained on Model2Model) does not do as well as ours. Thus, studio-trained Model2Model methods are insufficient for in-the-wild try-on, especially for limb reconstruction and garment warping.

\vspace{-3mm}
\subsubsection{Effectiveness of each component.}
\vspace{-2mm}
Fig.~\ref{fig:ablation_effectiveness} show an ablation study to verify the effectiveness of each component of our method.
We start with a baseline (A), in which we directly inpaint the masked-out regions of the person's original garment with a CLIP~\cite{radford2021clip} image embedding of the new garment. Then we add the following components one by one: (B) DensePose warping, (C) skin Inpainting, and (D) Warping Correction. 
The results show that directly generating the garments without warping cannot reconstruct the garment details (Fig.~\ref{fig:ablation_effectiveness} A). Without the proposed skin inpainting, the result will be biased toward the shape of the original garment (Fig.~\ref{fig:ablation_effectiveness} B). Lastly, without the warping correction, we cannot properly reconstruct high-frequency textures like stripes (Fig.~\ref{fig:ablation_effectiveness} C). 

Note that in Fig.~\ref{fig:ablation_effectiveness}-a, the FID discrepancy is not large between (C) and (D) when the person is from the street. The reason is that the complex backgrounds in the street images reduce the difference in data distributions, so FID does not clearly reflect the improvement in texture refinement. Instead, the effectiveness of the warping correction module can be clearly seen in the Shop2Model test, where the person images have a clean background so that the FID can focus on the garment details.

%% file: tables/person2person_table_simplified.tex
\begin{table}[!htb]
\centering
\vspace{-3mm}

\resizebox{\textwidth}{!}{
\begin{tabular}{|l||c|c|c|}
\hline
\multirow{5}*{}
     
     & Street2Street 
     & Model2Street
     & Model2Model
     \\ 
      
     & \small{FID $\downarrow$} 
     & \small{FID $\downarrow$}
     & \small{FID $\downarrow$}
     \\\hline
     

Ours (Paired, VITON-HD)
& 33.165
& \textbf{34.050} 
& 10.961
\\ 

Ours (Unpaired, VITON-HD)
& 33.742 & 34.434 & 11.040  \\ 

Ours (Unpaired, StreetTryOn)
& \textbf{33.039} & 34.191  & \textbf{10.214}
 \\
\hline

FS-VTON~\cite{he2022styleflow} 
& 67.009 
& 77.273
& 13.926
\\ 

HR-VITON~\cite{lee2022hrvton} 
& 63.539 
& 55.172
& 20.404
\\ 

SDAFN~\cite{bai2022sdafn} 
& 42.432 
& 44.537
& 14.316
\\

PWS~\cite{albahar2021posewithstyle} 
& 84.326
& 76.889
& 34.224
\\

PastaGAN++~\cite{xie2022pastagan++} 
& 67.016 
& 71.090
& 13.848
\\

PastaGAN++ (street)
& 67.088
& 70.461
& 40.841

\\ \hline

\end{tabular}
}

\vspace{-3mm}
\caption{\small
\textbf{Evaluation on Street2Street, Model2Street, and Model2Model tests.} 
We retrain FS-VTON, HR-VTON and SD-VTON on paired DeepFashion dataset for Model2Model try-on at 512$\times$320. 
PWS is trained on paired DeepFashion~\cite{liu2016deepfashion}, and PASTAGAN++ is trained on UPT dataset~\cite{xie2021pastagan}. PASTAGAN++ (street) is trained on the proposed Street TryOn dataset.\label{tab:person2person_table}
}

\end{table}


%% file: tables/shop2person_table_simplified.tex
\begin{table}[!htb]

\centering
\vspace{-5mm}
\resizebox{\textwidth}{!}{

\begin{tabular}{|l||c|ccc|}
\hline
\multirow{5}*{}
    
     & Shop2Street 
      & \multicolumn{3}{c|}{Shop2Model (VITON-HD)} \\ 
      
     & \small{FID $\downarrow$}
     & \small{FID $\downarrow$}
      & \small{SSIM $\uparrow$} 
      & \small{LPIPS $\downarrow$} 
     \\\hline


Ours (Paired, VITON-HD)
& \textbf{33.819} 
& 9.671 & 0.840 & 0.113

\\ 

Ours (Unpaired, VITON-HD)
&35.135 
& 11.675 & 0.826 & 0.128
 \\ 

Ours (Unpaired, StreetTryOn)
& 34.054 
&11.951 & 0.823 & 0.129
\\ \hline
     

SDAFN~\cite{bai2022sdafn} 
& 62.735
& 9.400 & 0.882 & 0.092

\\ 

FS-VTON~\cite{he2022styleflow} 
& 77.843
& 9.552 & 0.883 & 0.091

\\ 

HR-VITON~\cite{lee2022hrvton} 
& 63.516
& 16.21  & 0.862  & 0.109

\\

GP-VTON~\cite{xie2023gp-vton} 
& n.a.
 & 9.197 & 0.894 & 0.080

\\

StableVITON~\cite{kim2023stableviton} 
& 37.085
 & \textbf{8.233} & \textbf{0.888} & \textbf{0.073}

\\ \hline

\end{tabular}
}

\caption{
\textbf{\small Evaluation on Shop2Street and Shop2Model tests} 
 at 512$\times$320 and 512$\times$384 respectively. The methods are retrained at 512$\times$384 if their released models have a lower resolution. We resize output images to the resolution for these methods with released models at higher resolutions.
}
\vspace{-3mm}
\label{tab:shop2person_table}
\end{table}

%% file: tables/ablation_bg_removal.tex
\begin{figure}

\vspace{-5mm}
\centering
    \resizebox{0.8\textwidth}{!}{
        
        \begin{tabular}[b]{|c||c c c|}
        \hline
               & \makecell{FG-tryon only \\ (w/o BG)} & \makecell{FG-tryon + \\ BG inpainted} & \makecell{end-to-end \\ (w/ BG)}  \\ \hline
        FS-VTON     & 35.137     & 53.018    & 67.009   \\
        HR-VITON    & 41.154     & 55.857    & 63.539   \\
        SDAFN       & 34.274     & 47.634    & 42.436   \\
        PASTAGAN++  & 36.267      & 49.732   & 67.016 \\ \hline 
        Ours         & \textbf{31.218}  &     \textbf{46.178}   & \textbf{33.039}   \\
        \hline
        \end{tabular}
    }
    \includegraphics[width=0.8\textwidth]{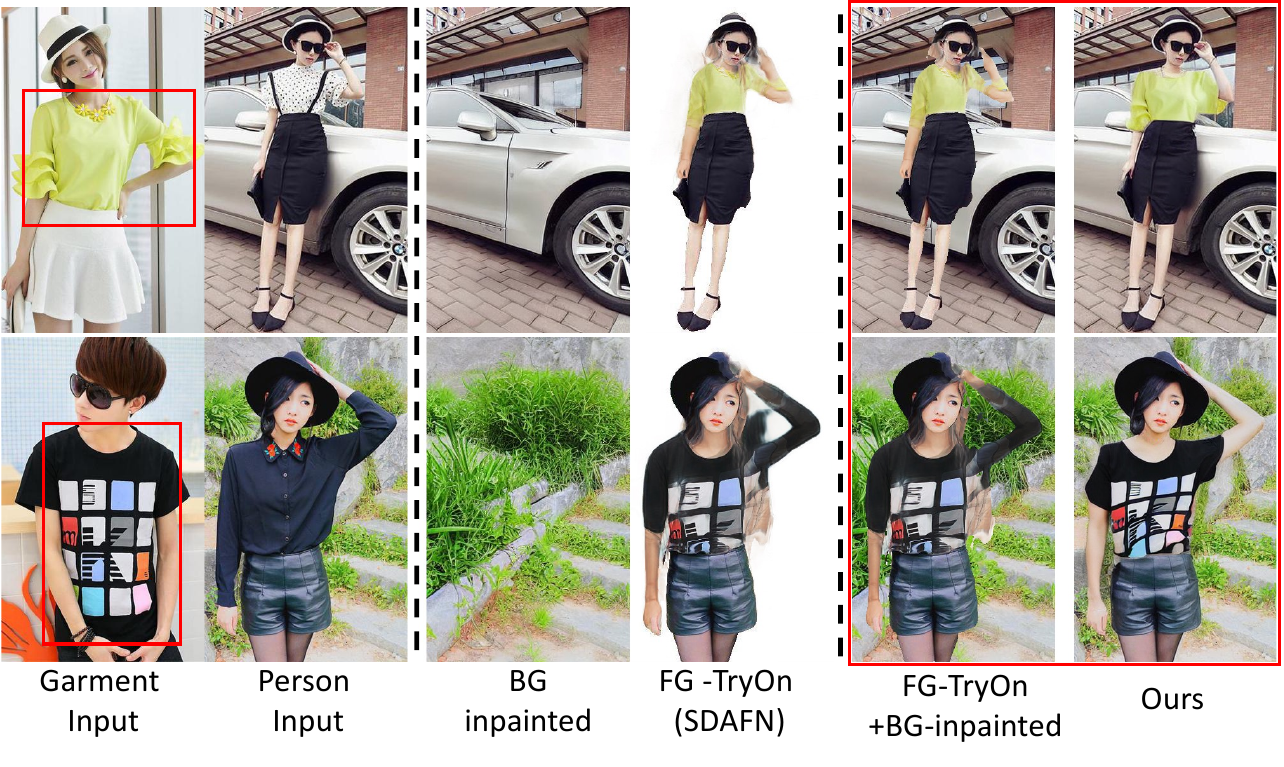}

\vspace{-5mm}
\caption{
The influence of image background in Street2Street try-on task. We report Street2Street try-on for segmented foreground try-on only (FG-tryon), foreground try-on blended with inpainted background (FG-tryon + BG inpainted), and running inference for each model without foreground segmentation (end-to-end). 
\textbf{(Top)}: Quantitative comparisons with FS-VTON, SDAFN, HR-VTON, and PASTAGAN++, which are all {\bf retrained} for Model2Model try-on, if any is originally not trained for another setting.
\textbf{(Bottom)}: Visual comparison with the best prior method, SDAFN.
  }\label{fig:bg_removal}

\end{figure}


%% file: tables/ablation_effectiveness.tex
\begin{figure*}
\centering

\vspace{-5mm}
\subfloat[\label{fig:visual_ablation_component}][Visual Results.]{
    \includegraphics[width=0.6\textwidth]{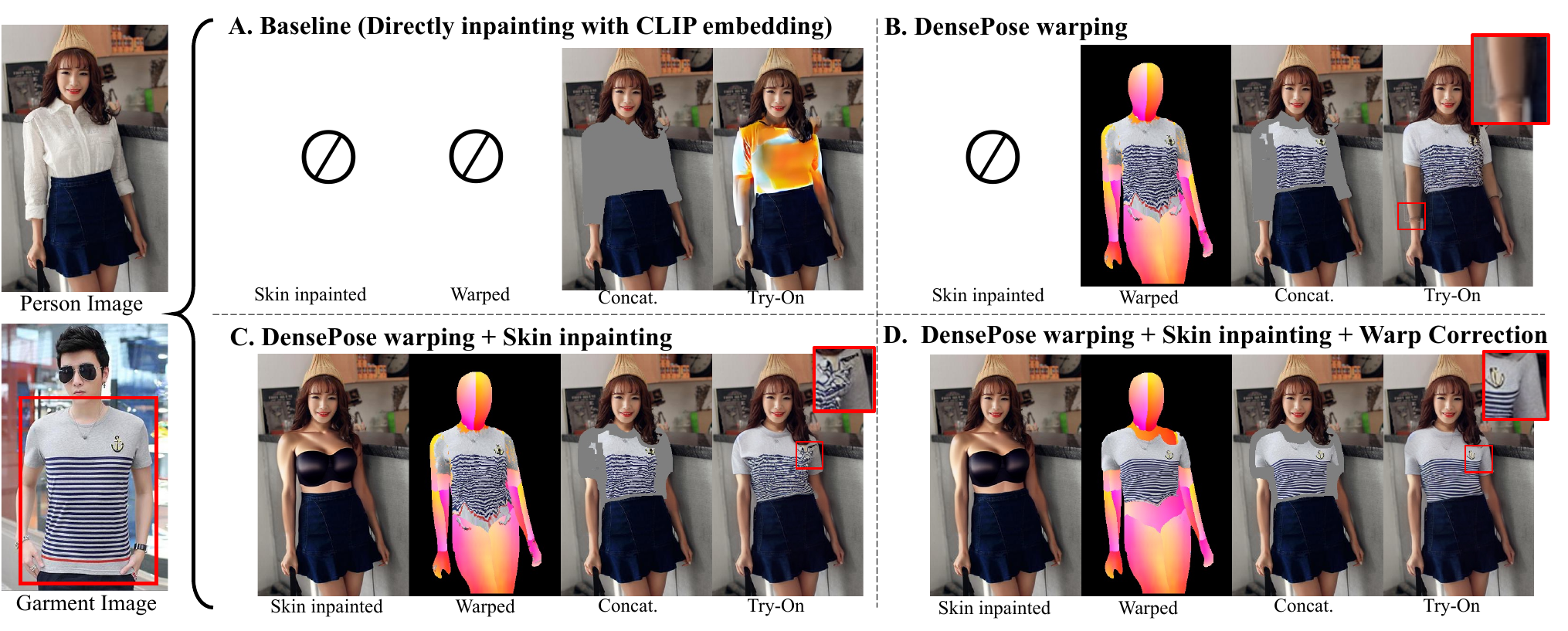}
}
\subfloat[\label{fig:table_ablation_component}][Quantitative Results (FID).]{
\resizebox{0.4\textwidth}{!}{
\begin{tabular}{|c||c|c|c|cc|}
\hline
\multirow{2}*{}

     & Shop2Shop
     & Shop2Street
     & Model2Street
     & \multicolumn{2}{c|}{Street2Street} 
     \\ 
     & \small{Top} 
     & \small{Top}  
     & \small{Top} 
     & \small{Dress} 
     & \small{Top}  
     \\\hline
     
A
& 54.482
& 62.247
& 59.732
& 100.587
& 59.665

\\\hline

B
& 22.942
& 36.319
& 34.412
& 38.974
& 34.853
\\\hline

C
& 14.024
& \textbf{35.368}
& 34.218
& \textbf{34.115}
& 34.009

\\\hline

D 
& \textbf{11.951}
& 35.632
& \textbf{34.191}
& 34.741
& \textbf{33.039}
\\\hline

\end{tabular}
}
\vspace{10mm}
}

\vspace{-3mm}
\caption{
Ablation study.
(A) Baseline that inpaints the masked-out original garment with CLIP vision embedding of the input garment; (B) adding DensePose warping; (C) adding skin inpainting on top of B; (D) adding warping correction on top of C. 
}
\label{fig:ablation_effectiveness}
\end{figure*}

%% file: sec/6_conclusion.tex
\section{Conclusion}

\begin{figure}
    \centering
    
    \vspace{-5mm}
    \includegraphics[width=0.7\textwidth]{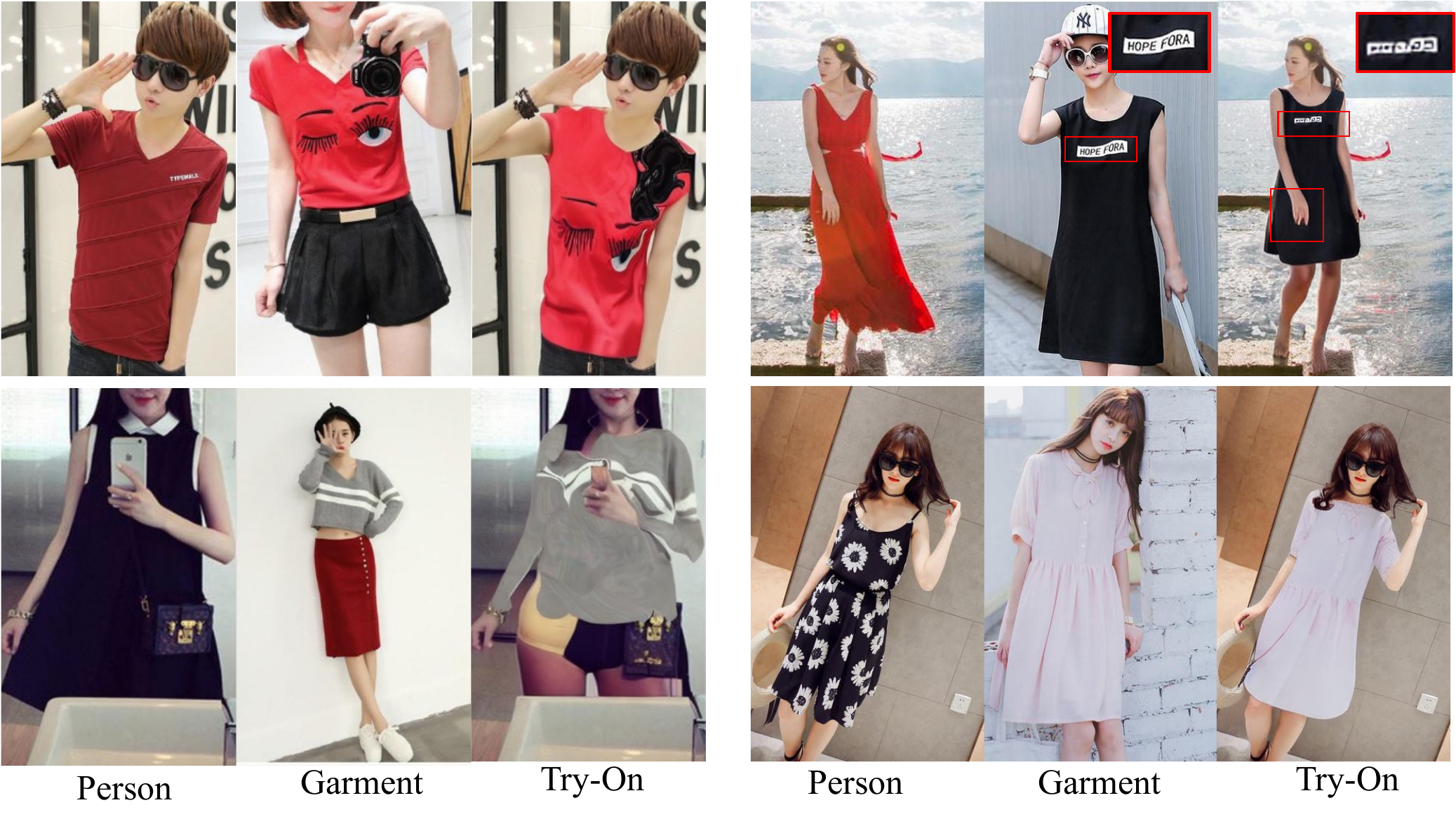}
    \vspace{-3mm}
    \caption{\small Failure cases. (Top-left) The camera is treated as part of the shirt. (Bottom-left) Warping failure. (Top-right) Failure to preserve garment details, hand artifact. (Bottom-right) Lighting mismatch, neck artifact.}
    \label{fig:failure_cases}
\end{figure}

In this work, we have introduced a new \textbf{StreetTryOn} benchmark and proposed a try-on method leveraging powerful pre-trained pose estimation and inpainting networks to robustly transfer garments to and from in-the-wild images. Although our method shows promising results, it still has room to improve, as shown by the failure cases in Fig.~\ref{fig:failure_cases}. In particular, our method cannot perform relighting, inherits artifacts like bad hand generation from Stable Diffusion, and does not always preserve garment details. 
Many of these issues can be alleviated with more fine-tuning and better correction modules.

\paragraph{Acknowledgements.}
We thank Zhenyu Xie, the first author of prior work GP-VTON~\cite{xie2023gp-vton} and PASTAGAN++~\cite{xie2022pastagan++}, for sharing the results of prior work and helping with implementations of prior work.

%% file: supp/dataset_filtering_details.tex
\section{Street TryOn Benchmark Details}

The proposed Street TryOn benchmark is derived from the Fashion Retrieval Dataset DeepFashion2~\cite{ge2019deepfashion2}. DeepFashion2 releases $191,961$ and $32,153$ in-the-wild fashion images for training and validation. These images feature models wearing an assortment of clothing items belonging to 13 popular clothing categories. For each image, a comprehensive set of annotations is available, encompassing information on scale, occlusion, zoom-in level, viewpoint, category, style, bounding box coordinates, dense landmarks, and per-pixel masks. 

However, DeepFashion2 images cannot be directly used for Virtual Try-On, which requires a frontal-view person with at least the upper body fully present and without large occlusion in relatively bright lighting conditions. Therefore, a two-stage data filtering process was employed for both the training and validation sets.

\subsection{Data Filtering Processes} 
\begin{figure}
    \centering
\includegraphics[width=0.7\textwidth]{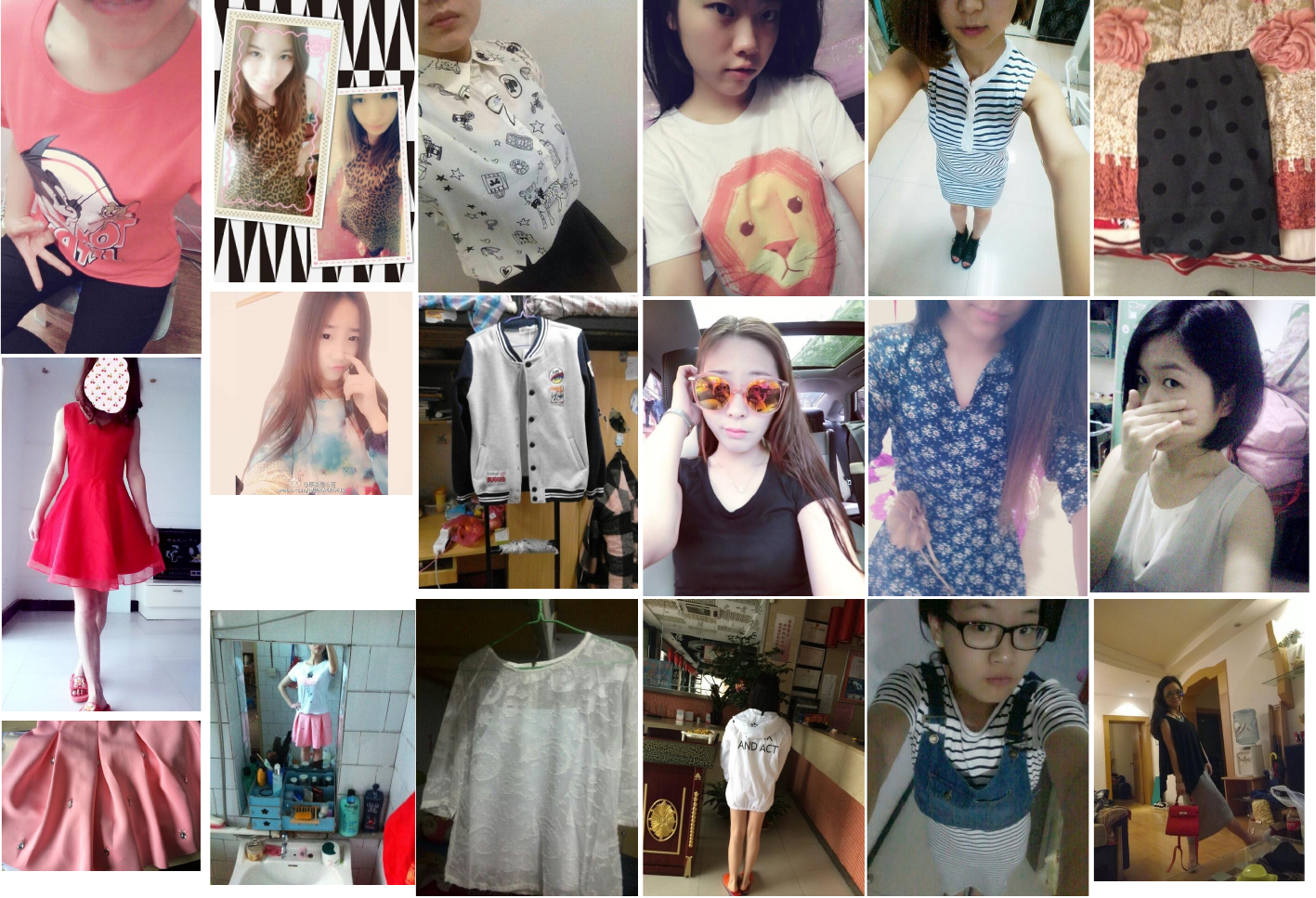}
    \caption{Example DeepFashion2 images that are not feasible for virtual try-on tasks and filtered out.}
    \label{fig:unselected_deepfashion2}
\end{figure}

At the first stage of filtering, we use DeepFashion2~\cite{ge2019deepfashion2}'s provided annotations to keep only the images with labels ``frontal viewpoint'', ``no zoom in'' and ``slight occlusion''. Besides, we filtered out the image sourced from customers, which contains a large number of selfies and images taken in dark rooms, as shown in Fig.~\ref{fig:unselected_deepfashion2}. After the initial filtering stage, we get $15,556$ and $2,401$ training and test images. 

Subsequently, in the second stage of filtering, the focus was on identifying images that portrayed the entire upper body of the models. 
We run the DensePose detection\cite{guler2018densepose}, which also detects human bounding boxes on these images. We discarded images without human bounding boxes detected and images with the person present horizontally (e.g., lying down) or images with bounding boxes in an aspect ratio larger than $5:8$. We then pad the human bounding box to make its aspect ratio $5:16$, cropped the person from the images, and resized it to $512 \times 320$. 

\begin{figure}
    \centering
    \includegraphics[width=0.7\textwidth]{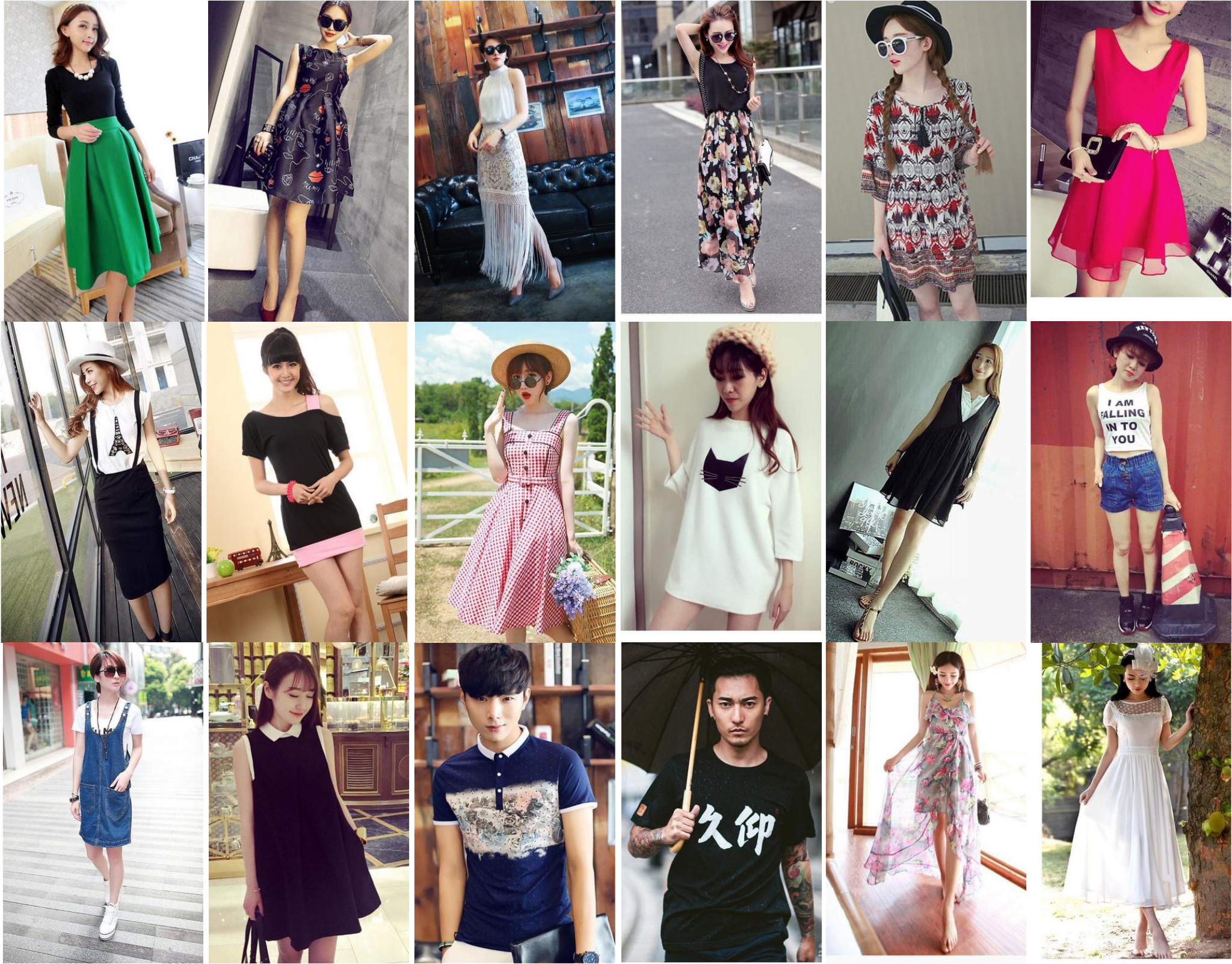}
    \vspace{-3mm}
    \caption{Examples of the selected DeepFashion2 images after cropping, which are included in the proposed Street TryOn Benchmark.}
    \label{fig:selected_deepfashion2}
\end{figure}

Following these two stages of meticulous filtering, the dataset was refined to encompass $12,364$ images for the training set and $2,089$ for the validation set. The examples of the selected and processed images can be found in Fig.~\ref{fig:selected_deepfashion2}.

\subsection{Data Statistics}
\begin{figure}
    \centering
    \includegraphics[width=0.8\textwidth]
    {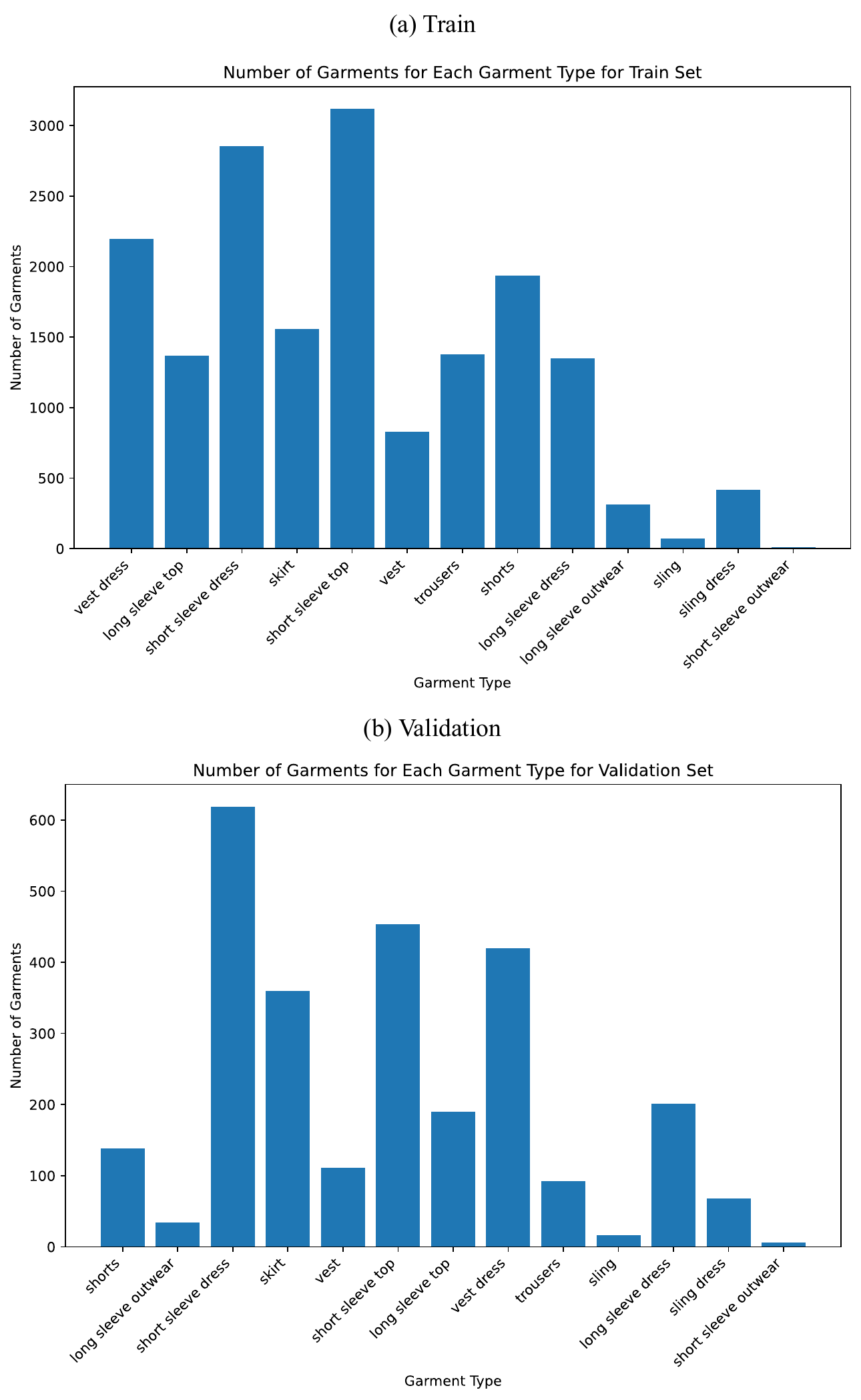}
    \vspace{-3mm}
    \caption{The number of garments in each garment type.}
    \label{fig:stats_num_category}
\end{figure}

 Because each image in the DeepFashion2 dataset has rich annotation provided, we further investigate the annotations of the Street TryOn benchmark derived from DeepFashion2, and we obtain a detailed data analysis for the data distributions. As shown in Fig.~\ref{fig:stats_num_category}, the proposed Street TryOn benchmark contains a diverse set of garments in various categories.

\begin{figure}
\centering
\includegraphics[width=0.8\textwidth]{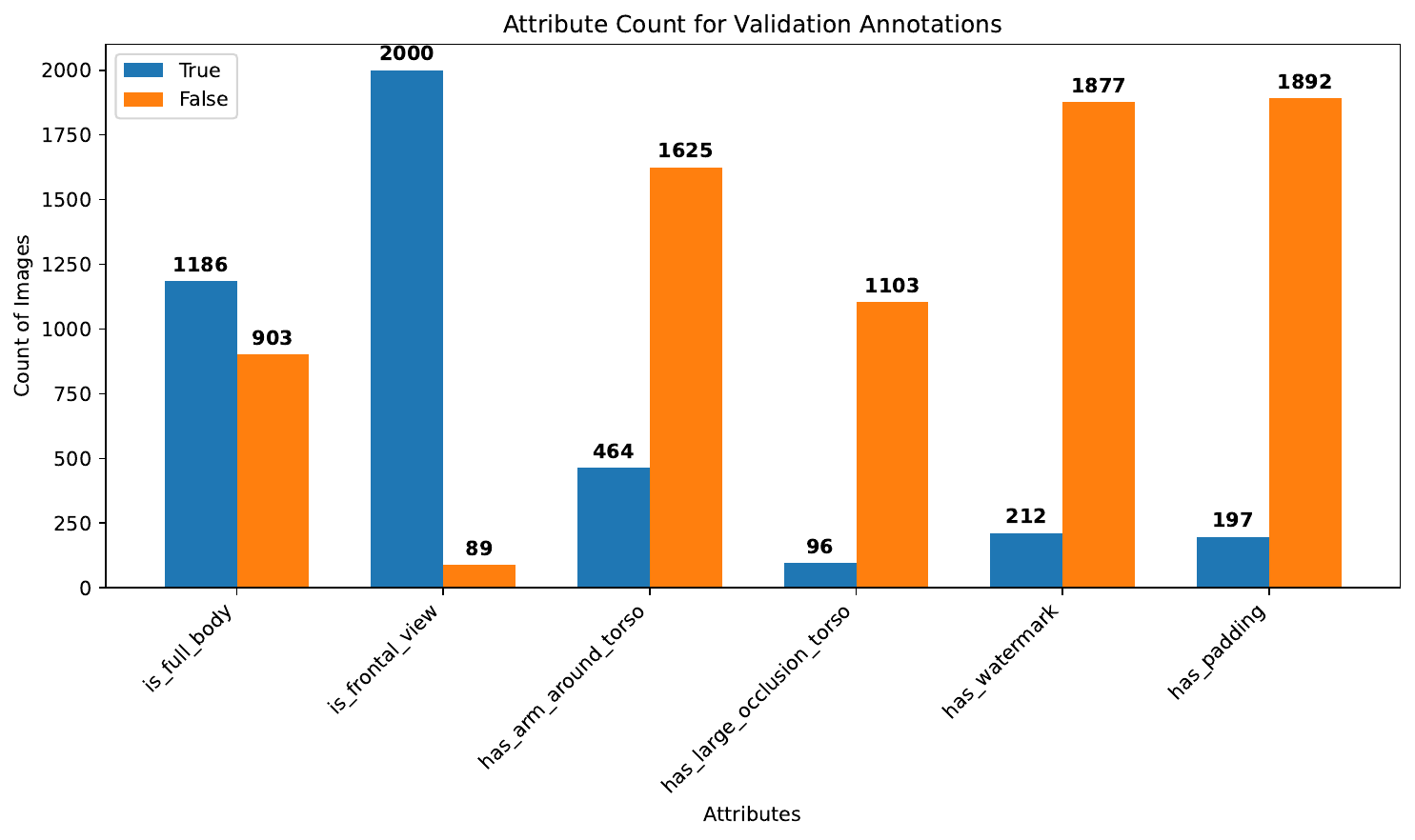}
\caption{The distribution of manually labeled attributes in the test set.}
    \vspace{-3mm}
\label{fig:stats_test_attribute}
\end{figure}

Next, we looked into the images in the test set and manually labeled six attributes for each image, which are ``is\_full\_body'', ``is\_frontal\_view'', ``has\_arm\_around\_torso'', ``has\_large\_occlusion\_torso'', ``has\_watermark'' and ``has\_padding''. These labels can be used to divide the validation set into different difficulty levels in the future. The attribute distributions can be found in Fig.~\ref{fig:stats_test_attribute}.

\begin{figure}
\centering
\includegraphics[width=0.8\textwidth]{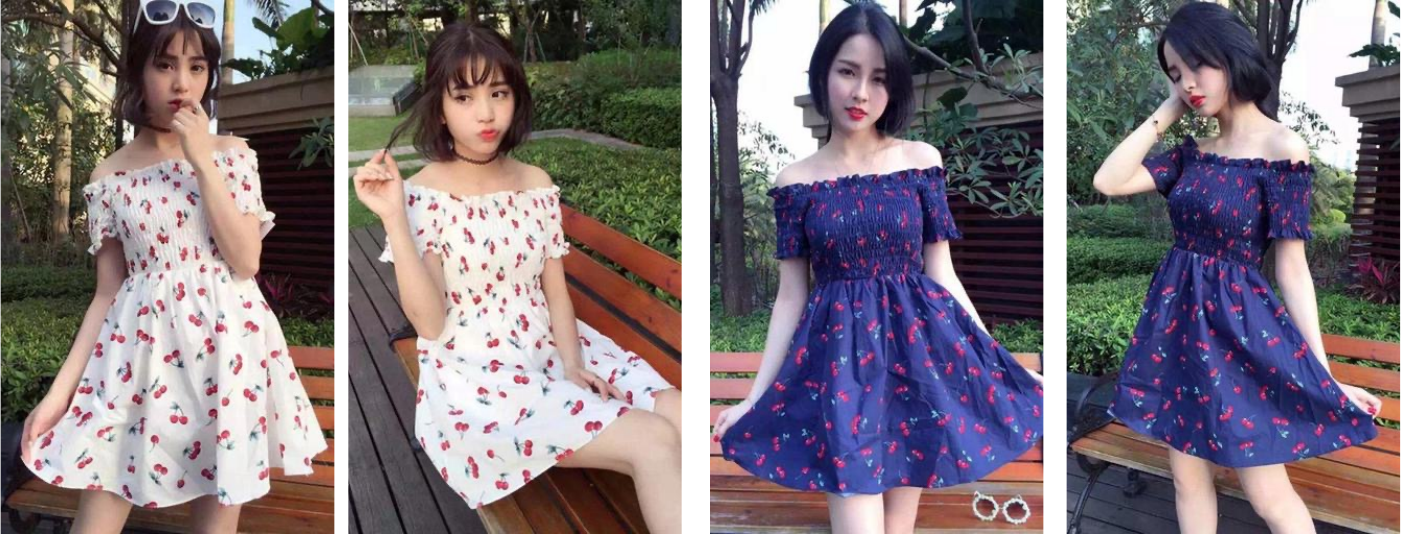}
\caption{The images that share the same     `garment\_id' in the DeepFashion2 Dataset. Color variations cannot be distinguished.}
\label{fig:garment_id}
\end{figure}

Since DeepFashion2 is a fashion retrieval dataset that originally contains multiple images for the same garments (by sharing the same `garment\_id'), we further investigate if we can build image pairs to enable paired training and validation for virtual try-on tasks. 
After the filtering processes, in the Street TryOn benchmark, there are 14\% unique images without any other images labeled as the same garments. For the rest of the images, although the DeepFashion2 annotation suggests they have potentially paired images, the annotation is too noisy to be directly used to construct pairs, because the color variations of garments are not distinguished by the DeepFashion2's garment\_ids, as shown in Fig.~\ref{fig:garment_id}. Therefore, if one wants to build paired images for the Street TryOn benchmark in the future, additional manual annotation would be necessary.

%% file: supp/cosine_noise_details.tex
\section{DensePose Perturbation with Cosine Noise}
\begin{figure*}
    \centering
    \includegraphics[width=\textwidth]{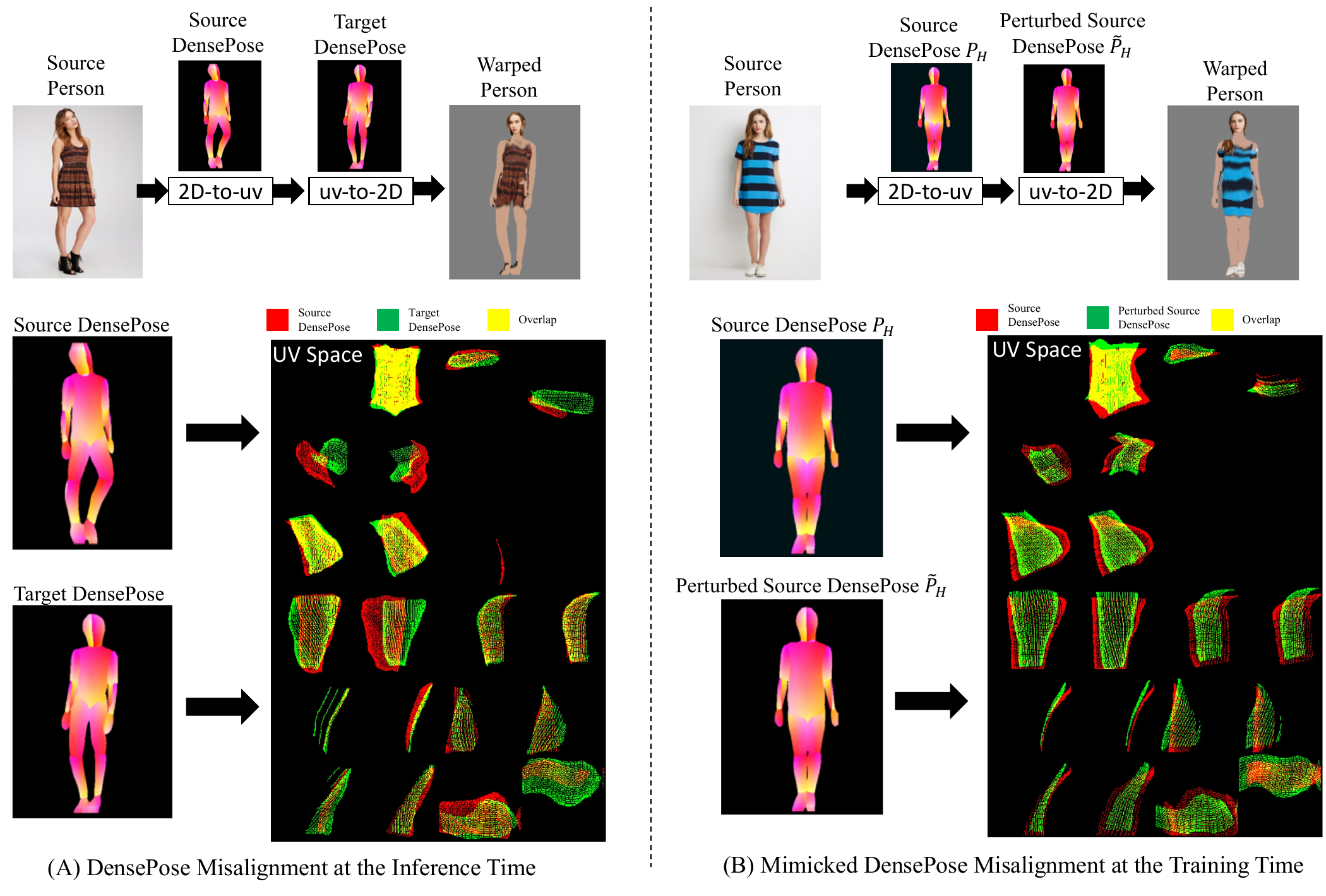}
    \vspace{-3mm}
    \caption{\textbf{DensePose Perturbation with Cosine Noise.} The example full-body person images are from the DeepFashion dataset~\cite{liu2016deepfashion} and are only used in this figure for illustration purposes. In this illustration, the full person is warped to demonstrate how the 24 UV maps in DensePose (for 24 different body parts) are misaligned. In the final warped image, we fill the skin with the average skin color for visualization.}
    \label{fig:cosine_perturbation}
\end{figure*}
As mentioned in the main paper, when we train the DensePose correction module, a cosine noise is added to the DensePose's pixel values to mimic the imperfect DensePose alignment at the test time. This way, we can train it without paired data but still achieve robust performance at the test time.

Fig.~\ref{fig:cosine_perturbation}(A) shows how the imperfect DensePose prediction causes warping distortion when we transform a garment from one pose to another. Clearly, in the visualized UV overlapping, the source DensePose and the target DensePose are not perfectly matched.

After observing the misalignment patterns, we propose to use a cosine perturbation to mimic this misalignment. Given a person's DensePose $P_H$, in which each pixel $(i,j)$ contains a $(u,v)$ coordinate to map the person into UV space as $P_H[i,j] = (u,v)$, we add a cosine noise to its pixel values as 
\begin{equation}
    \tilde{P}_H[i,j] = \Bigl(
    u + k_1 \cos(\alpha_1 u + \beta_1),
    v + k_2 \cos(\alpha_2 v + \beta_2) \Bigr), 
\end{equation}
where $k_1,k_2,\alpha_1, \alpha_2, \beta_1$ and $\beta_2$ are randomly sampled coefficients for each UV map in DensePose $P_H$. 
As shown in Fig.~\ref{fig:cosine_perturbation}(B), the cosine perturbation can effectively simulate the DensePose misalignment at the inference time. Therefore, with the cosine perturbation, we can mimic the inference scenarios during the training time without paired data.

%% file: supp/diffusion_inpainting_details.tex
\begin{figure}
\centering
\includegraphics[width=\textwidth]{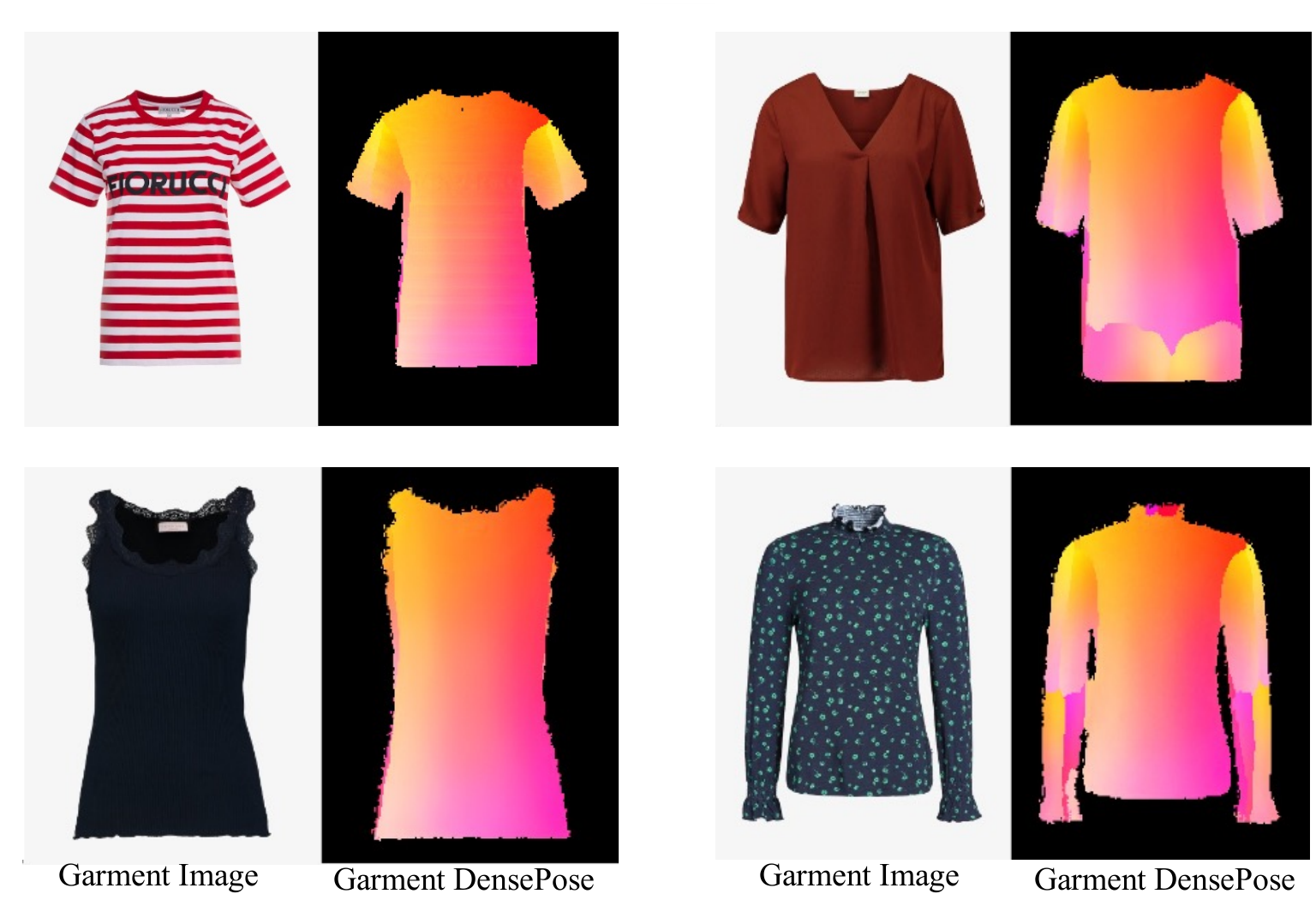}
\caption{The detected garment DensePose.}
\label{fig:garment_densepose}
\end{figure}
\section{Additional Implementation Details}

\subsection{Pre-trained Diffusion Models}
Here, we provide additional implementation details of the diffusion models used in this work. 

We set the inference steps for both the skin and refinement diffusion inpainters as $20$. The guidance scale for both is $7.5$. The skin inpainter has the negative prompt set as ``art, clothes, garments, long-sleeves, sleeves, cloak, loose, thick clothes, loose clothes, pants, shirts, skirts, dresses, long jackets, jackets, cloth between legs, cloth around the body, cloth around arms.'' 
The refinement diffusion inpainter uses a negative prompt as ``blurry, cracks on skins, poor shirts, poor pants, strange holes, bad legs, missing legs, bad arms, missing arms, bad anatomy, poorly drawn face, bad face, fused face, cloned face, worst face, three crus, extra crus, fused crus, worst feet, three feet, fused feet, fused thigh, three thighs, fused thigh, extra thigh, worst thigh, missing fingers, extra fingers, ugly fingers, long fingers, horn, extra eyes, huge eyes, 2girl, amputation, disconnected limbs, cartoon, cg, 3d, unreal, animate''.

As defined by the pretrained diffusion inpainter~\cite{stable-diffusion-inpaint}, at the inference time, it takes a concatenation of noise, the latent masked image, and the mask as the input to start denoising.
Instead of directly using the Gaussian noise, our noise is built by first broadcasting the mean features into each segmentation class for the latent masked image based on the predicted tryon human parse $M_T$  and adding noises to it with timestep as $999$.

\subsection{Garment DensePose Detection}
For preprocessing, we implement and train the garment DensePose Detection algorithm proposed in the prior work of Cui et al.~\cite{cui2023learning} at resolution $256\times192$. Then, we resize the detected DensePose to the desired resolutions. The examples of detected garment DensePose can be found in Fig.~\ref{fig:garment_densepose}.

%% file: supp/more_results.tex
\section{More Visual Results}
In this section, more visual results are presented to verify the effectiveness of our approach.

\subsection{Shop2Model Test}
\begin{figure*}
    \centering
    \includegraphics[width=0.95\textwidth]{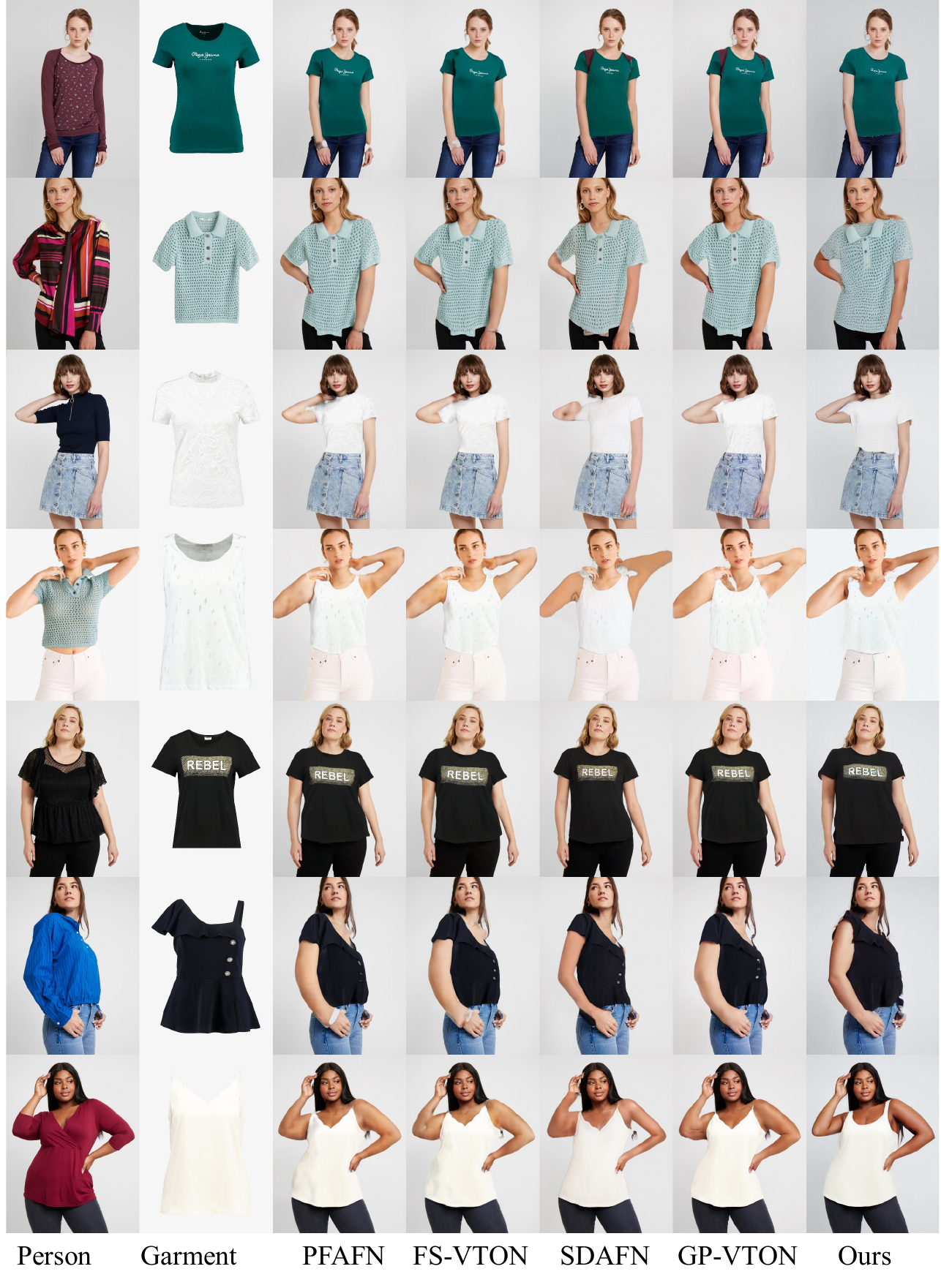}
    \vspace{-3mm}
    \caption{\textbf{More examples for Shop2Model test} on VITON-HD benchmark~\cite{choi2021vitonhd} 1.}
    \label{fig:vitonhd_batch0}
\end{figure*}

\begin{figure*}
    \centering
    \includegraphics[width=0.95\textwidth]{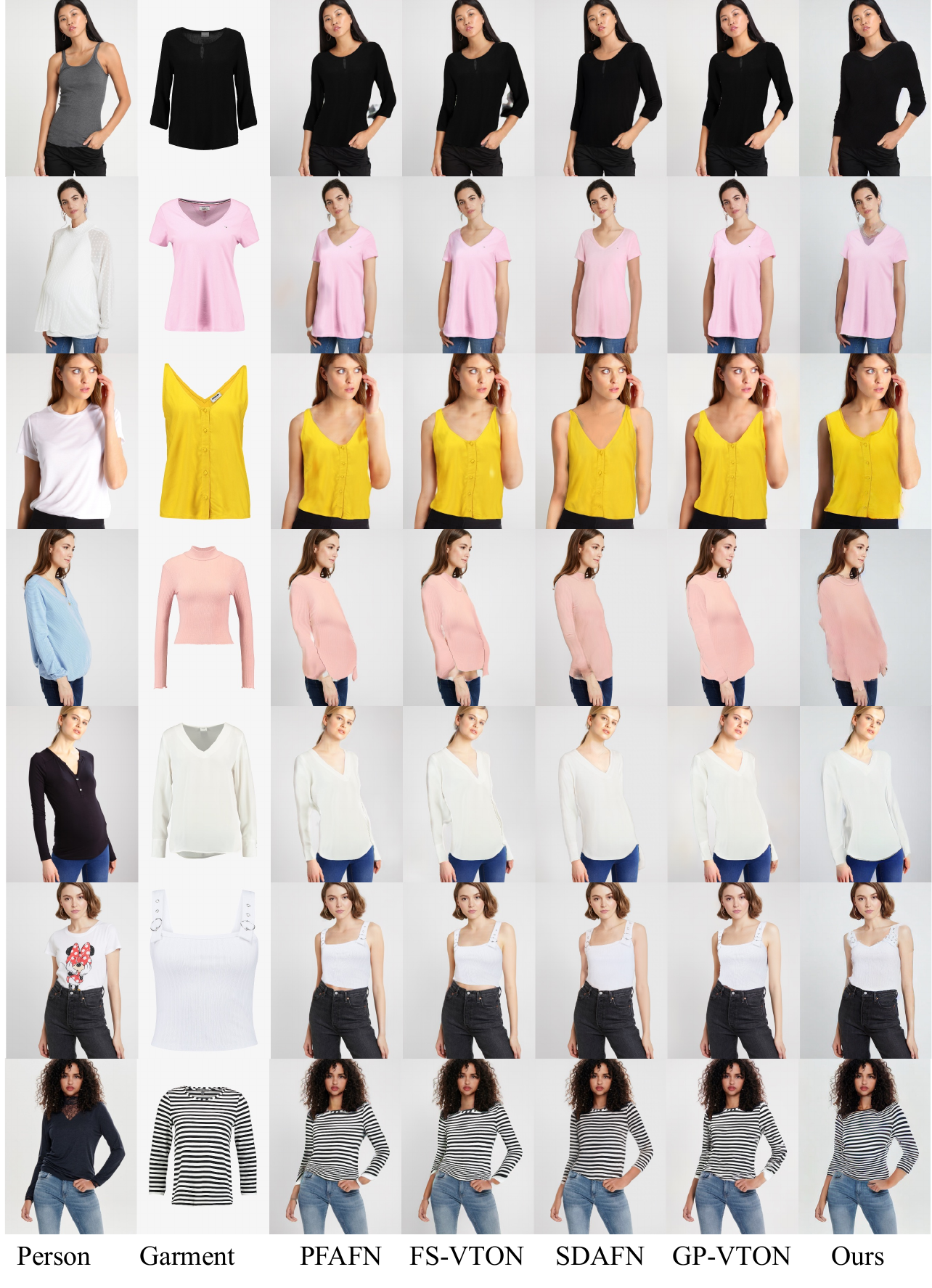}
    \vspace{-3mm}
    \caption{\textbf{More examples for Shop2Model test} on VITON-HD benchmark~\cite{choi2021vitonhd} 2.}
    \label{fig:vitonhd_batch1}
\end{figure*}

More results can be found in the Shop2Model test on VITON-HD in Fig~\ref{fig:vitonhd_batch0} and Fig.~\ref{fig:vitonhd_batch1}, comparing with PFAFN~\cite{ge2021pfafn}, FS-VTON~\cite{he2022styleflow}, SDAFN~\cite{bai2022sdafn} and GP-VTON~\cite{xie2023gp-vton}.

\subsection{Street2Street Test}
\begin{figure*}
    \centering
    \includegraphics[width=0.92\textwidth]{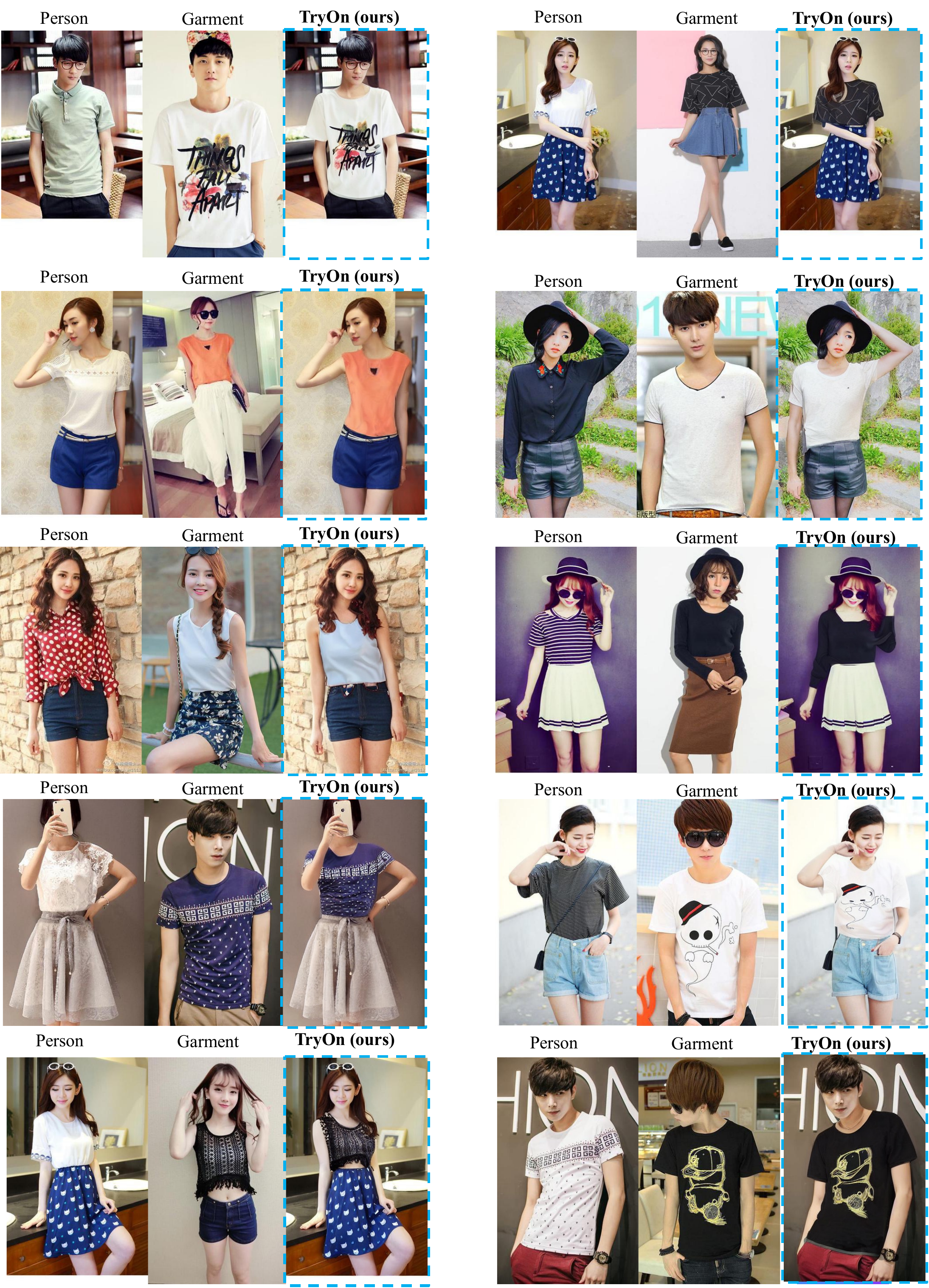}
    \vspace{-3mm}
    \caption{\textbf{More examples for Street2Street test} for top try-on.}
    \label{fig:street2street_top}
\end{figure*}

\begin{figure*}
    \centering
    \includegraphics[width=0.92\textwidth]{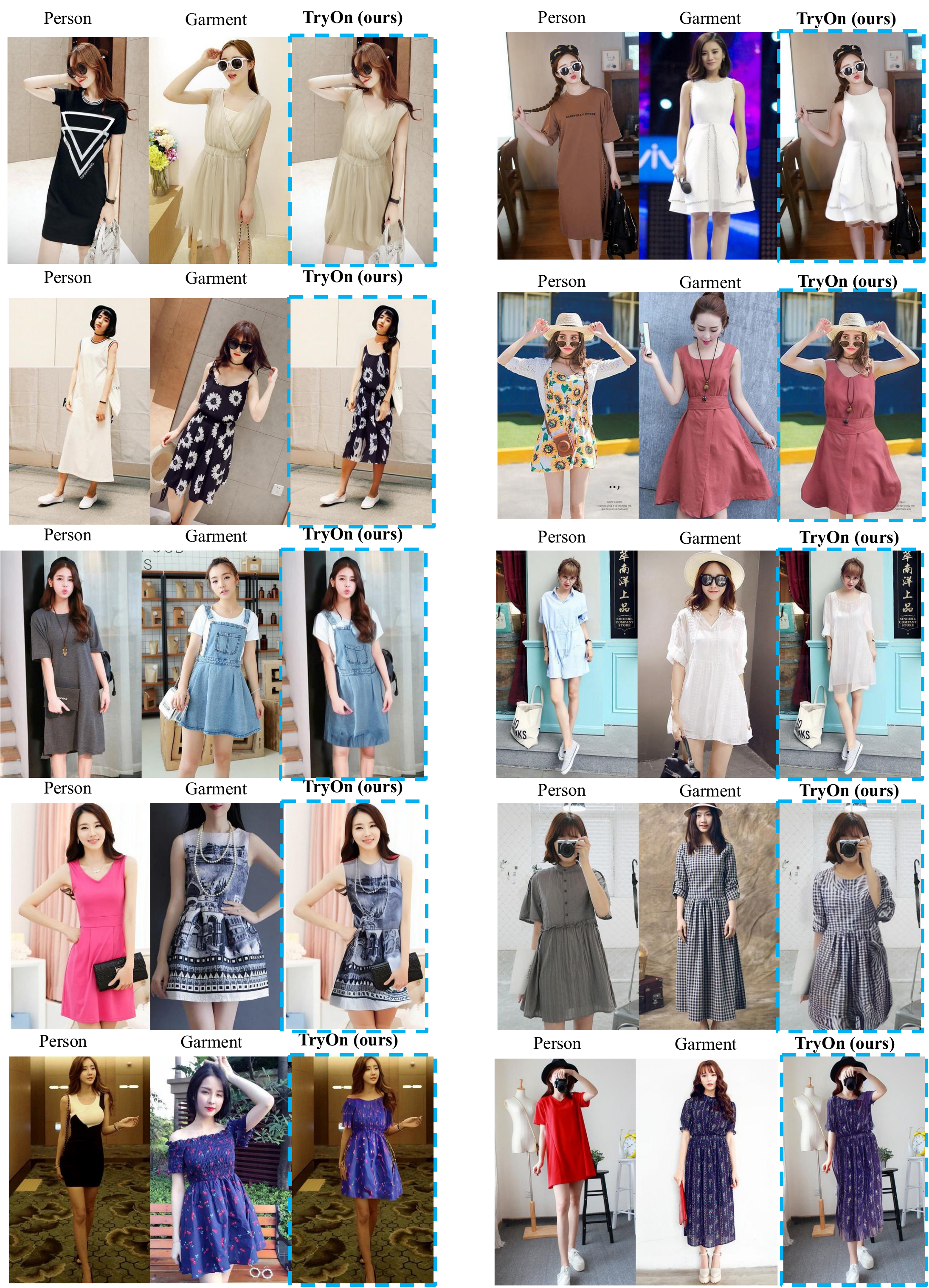}
    \vspace{-3mm}
    \caption{\textbf{More examples for Street2Street test} for dress try-on. }
    \label{fig:street2street_dress}
\end{figure*}

We finally show more results for our Street2Street try-on with intermediate outputs in Fig.~\ref{fig:street2street_top} and Fig.~\ref{fig:street2street_dress}.  

\subsection{Ablation Study}
\begin{figure*}
    \centering
    \includegraphics[width=\textwidth]{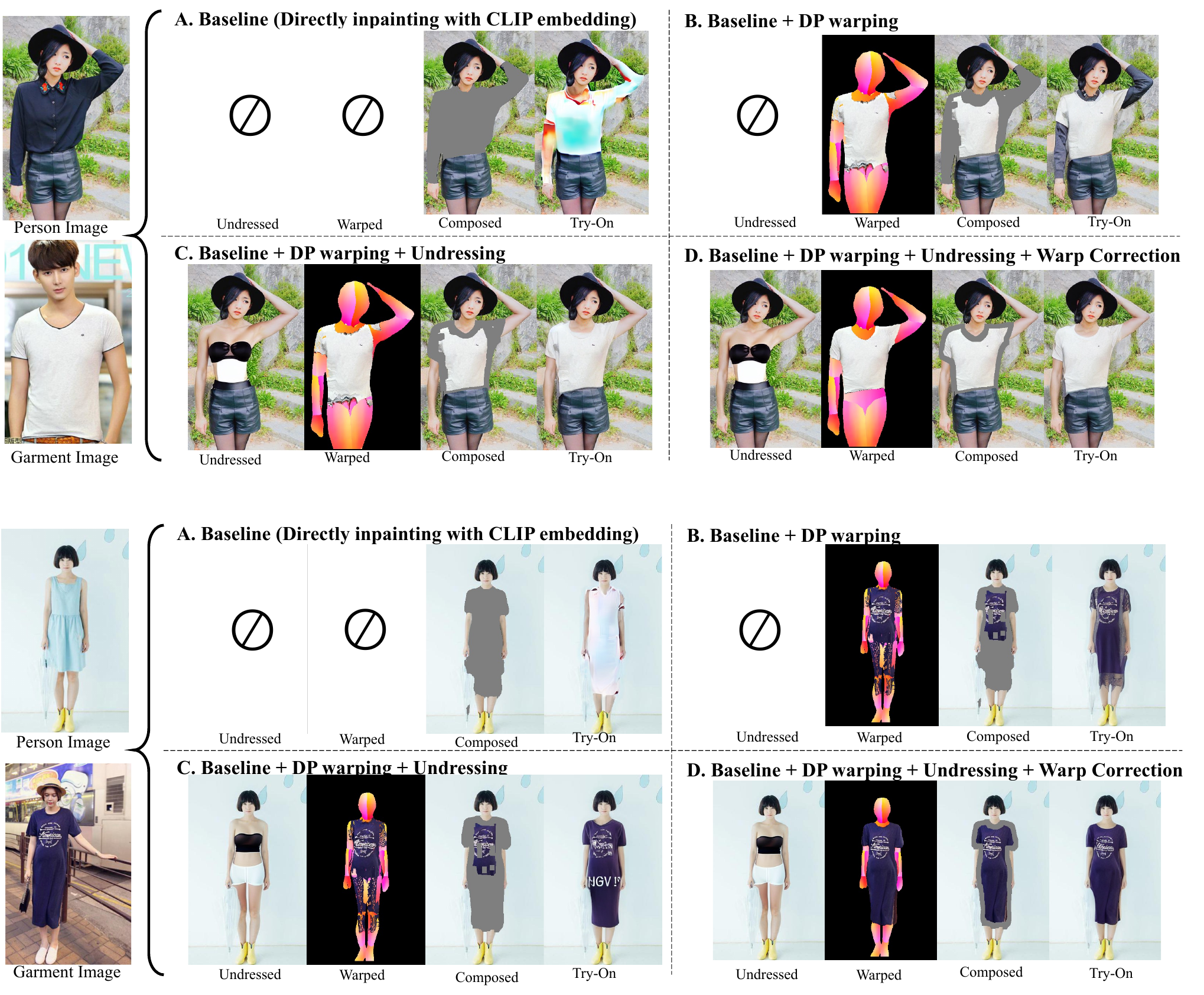}
    \vspace{-3mm}
    \caption{More examples for the ablation study to verify the effectiveness of each component in the proposed method. }
    \label{fig:more_ablation_examples}
\end{figure*}

Here, we also provide more examples of ablation studies to validate the effectiveness of each component in the design of our approach in Fig.~\ref{fig:more_ablation_examples}.
